\documentclass[review]{elsarticle}

\usepackage{framed,multirow}
\usepackage{floatrow}
\usepackage{amssymb}
\usepackage{latexsym}

\usepackage{url}
\usepackage{xcolor}
\usepackage{hyperref}
\usepackage{makecell}
\usepackage{lineno,hyperref,color}
\usepackage[figuresright]{rotating}
\usepackage{makecell}
\usepackage[ruled,vlined,linesnumbered]{algorithm2e}

\usepackage{amsmath}

\definecolor{newcolor}{rgb}{.8,.349,.1}

\usepackage{lineno,hyperref}
\modulolinenumbers[5]

\modulolinenumbers[5]

\DeclareMathOperator*{\argmin}{arg\,min}

\journal{Information Sciences}

\begin{document}

\newcommand\KH[1]{\textcolor{black}{#1}}
\newcommand\KHC[1]{\textcolor{blue}{#1}}

\begin{frontmatter}

\title{ASMCNN: An Efficient Brain Extraction Using Active Shape Model and Convolutional Neural Networks}%
\author[firstaddress]{Duy H. M. Nguyen}
\author[secondaddress]{Duy M. Nguyen}
\author[thirdaddress]{Mai T. N. Truong}
\author[fourthaddress]{Thu Nguyen}
\author[fifthaddress]{Khanh T. Tran}
\author[sixthaddress]{Nguyen A. Triet}
\author[seventhaddress]{Pham T. Bao}
\author[eighthaddress,ninthaddress]{Binh T. Nguyen\corref{correspondingauthor}}
\cortext[correspondingauthor]{Corresponding author}
\ead{ngtbinh@hcmus.edu.vn}
\address[firstaddress]{Max Planck Institute for Informatics, Germany}
\address[secondaddress]{Dublin City University, Ireland}
\address[thirdaddress]{Department of Multimedia Engineering, Dongguk University, South Korea}
\address[fourthaddress]{Department of Mathematics, University of Louisiana at Lafayette, USA}
\address[fifthaddress]{Center for Machine Vision and Signal Analysis, University of Oulu, Finland}
\address[sixthaddress]{Department of Mathematics,  University of Architecture of Ho Chi Minh City, Vietnam}
\address[seventhaddress]{Department of Computer Science, Saigon University, Ho Chi Minh City, Vietnam}
\address[eighthaddress]{Department of Computer Science, VNU-HCMUS, Ho Chi Minh City, Vietnam}
\address[ninthaddress]{Vietnam National University, Ho Chi Minh City, Vietnam}

\begin{abstract}
Brain extraction (skull stripping) is a challenging problem in neuroimaging. It is due to the variability in conditions from data acquisition or abnormalities in images, making brain morphology and intensity characteristics changeable and complicated. In this paper, we propose an algorithm for skull stripping in Magnetic Resonance Imaging (MRI) scans, namely ASMCNN, by combining the Active Shape Model (ASM) and Convolutional Neural Network (CNN) for taking full of their advantages to achieve remarkable results. Instead of working with 3D structures, we process 2D image sequences in the sagittal plane.
First, we divide images into different groups such that, in each group, shapes and structures of brain boundaries have similar appearances. Second, a modified version of ASM is used to detect brain boundaries by utilizing prior knowledge of each group. Finally, CNN and post-processing methods, including Conditional Random Field (CRF), Gaussian processes, and several special rules are applied to refine the segmentation contours. Experimental results show that our proposed method outperforms current state-of-the-art algorithms by a significant margin in all experiments.
\end{abstract}

\end{frontmatter}

\linenumbers

\section{Introduction}
Whole brain segmentation is the problem of extracting brain regions from volumetric data, such as Magnetic Resonance Imaging (MRI) or Computed Tomography (CT) scans. It resulted in segmentation maps indicating \KH{parts of brain} after removing non-brain tissues such as eyes, fat, bone, marrow, and dura. Brain extraction is the \KH{prior} step in most neuroimaging analysis systems, which usually \KH{consists} of brain tissue classification and volumetric measurement \citep{Jiang2013}, template construction \citep{Maldjian2014}, and cortical and sub-cortical surface analysis \citep{Dale1999}. Early preprocessing steps such as the bias field correction can also benefit from brain extraction \citep{Sharma2010}. Therefore, brain extraction needs to be considered before producing high-performance methods.

Automatic skull stripping reduces processing time and prevents any human bias in the final results, especially in large-scale studies, where thousands of images with different characteristics and significant anatomical variations need to be examined. Most skull-stripping methods are optimized and validated for MRI T1-weighted images since high-resolution T1-weighted structural images are prevalent in clinical studies \citep{Kalavathi2016}, and they provide excellent contrast between different brain tissues. Existing brain extraction methods can be categorized as edge-based, template-based, label fusion with atlas-based, non-local patch-based \citep{Roy2017}, and recently, deep learning-based approaches.

Edge-based methods focus on detecting edges between the brain region and the non-brain region by considering differences in these structures' appearance. Several techniques have been employed, such as watershed \citep{Richard2013}, level set \citep{jinyoung2011}, histogram analysis \citep{Balan2012}, etc. Although these methods have proven their effectiveness and achieved comparable results, their performance tends to be less accurate when working with pathology, different sites, scanners, and imaging acquisition protocols \cite{sharma2010automated}.

Template-based methods register a subject to a template via affine transformation \citep{Iglesias2011} or deformable models \citep{wang2011} to create an initial estimation for the brain mask. After that, the boundary of the brain mask is segmented again by one classifier to increase the accuracy of the final results. Templates can involve one or more distinctive atlases. Therefore they are robust, stable in different conditions, and highly accurate. 

Label fusion with atlas-based techniques, such as Multi-Atlas Propagation and Segmentation (MAPS) \citep{Leung2011}, Advanced Normalization Tools (ANTs) \citep{avants2014}, and Pincram \citep{heckemann2015}, implement registration of multiple atlases to a target subject by using deformable models. After being registered to the target space, one can combine brain masks in all atlases using Simultaneous Truth And Performance Level Estimation (STAPLE) \citep{warfield2004} or joint label fusion. Since these approaches' primary operations are registration processes, their performance depends on the accuracy of the registration ones and the quality of the brain mask in each atlas. Also, the variability representation in brain anatomy usually requires a large number of atlases. Hence these methods are typically time-consuming and computationally intensive.

Non-local patch-based methods first transform atlases to subject spaces using appropriate affine registrations to estimate the subject's initial brain mask. A neighborhood searching process for each small patch, which is around the initial estimation of the brain boundary, is performed. Finally, patches are associated together, and similarity weights can be computed to generate the final brain mask. Several methods, such as Brain Extraction using non-local Segmentation Technique (BEAST) \citep{Eskildsen2012}, Multi-cONtrast brain STRipping (MONSTR) \citep{Roy2017} are inspired by this idea, and they achieve remarkable performance in aspects of both accuracy and robustness. However, one difficulty of this approach is pinpointing the optimal setting for parameters, including the number of atlases or the window size.
Additionally, in T1-weighted atlases, the intensity values of structures such as hemorrhages, tumors, or lesions may be similar to that of non-brain tissues. Non-local based methods use different MRI acquisition protocols to obtain complementary information and overcome this challenge. As a result, it is a complex task and requires more processing time. 

Recently, deep learning has become a hot trend in many fields, including neuroimaging. Unlike traditional methods, deep learning is more reliable, robust, and able to produce superior performance in pattern recognition. One of the most popular systems using a deep-learning approach in neuroimaging is U-Net \cite{ronneberger2015u}. In this method, one contracting path and another symmetric expanding path are used for capturing context and precise localization, respectively. U-Net can use available annotated samples more efficiently by utilizing data augmentation for the training phase. Kleesiek and colleagues \citep{kleesiek2016deep} propose a 3D convolutional deep learning architecture for skull stripping. This method can deal with pathologically altered tissues and different modalities; hence it can be applied to several types of MRI data. Hao Dong and collaborators \cite{dong2017automatic} present a fully automatic scheme that can detect and segment brain tumors simultaneously using the U-Net based deep convolution networks. The experiments on the BRATS 2015 challenging \cite{menze:hal-00935640} demonstrated that their method could provide both efficient and robust segmentation compared to the manual delineated ground truth. Other studies can be found at \cite{ORTIZ2014117,BINCZYK2017235}.

In this research, we present a novel approach for brain extraction in T1-weighted MRI data by combining ASM \citep{Cootes1995} and CNN \citep{Lecun1998}, namely ASMCNN. Unlike existing methods, we consider the brain extraction problem as a segmentation task for 2D image sequences in the sagittal plane rather than working with 3D structures. Our approach has several benefits. First, it allows us to develop specific algorithms for different brain boundary silhouettes. Throughout image sequences, shapes of brain boundaries and brain sizes vary significantly, especially for sagittal slices located at the beginning and ending parts of 3D MRI volumes, where brain regions are small, and the corresponding boundaries are very complex. Based on prior knowledge about brain structures \cite{netter2014atlas}, we develop specific rules to control the segmentation process effectively. Second, images in the sagittal plane are symmetrical across two hemispheres. By utilizing this property, we can predict the brain mask's general shape based on slices' positions. This property also enables us to establish more extensive and accurate rules for segmentation.

The main contributions of our work can be summarized as follows:
\begin{enumerate}
\item Our approach is motivated by the divide-and-conquer idea; thereby, we group 2D-images into different groups by exploiting the brain's symmetry in the sagittal plane. Each group is then designed to obtain optimal predictions that satisfy both global, local information, and prior medical knowledge.

\item CNN, with high-level feature representations, is utilized to refine brain boundaries. Besides, global spatial information of each voxel is combined with features computed from convolutional neural networks. Hence, our method can represent both global and local features simultaneously.
\item We make use of the similarity of brain contours in each group by applying ASM to detect general properties of boundaries. This procedure guarantees each object's geometrical attributes and enhances the overall accuracy of segmentation results.
\item Our framework does not depend on any specific MRI format, and therefore, it can be applied effectively in different acquisition conditions.
\end{enumerate}

The \KH{remainings} of this paper are organized as follows. In Section 2, we introduce all necessary notations and present an appropriate architecture and algorithms for the problem. After that, we compare our approach with seven state-of-the-art methods on three public datasets in Section 3.  The Section 4 is devoted to discuss strengths and major limitations of ASMCNN as well as future  research directions. Finally, we end the paper by a brief conclusion.
\begin{figure}[!t]
	\includegraphics[width=1.0\textwidth]{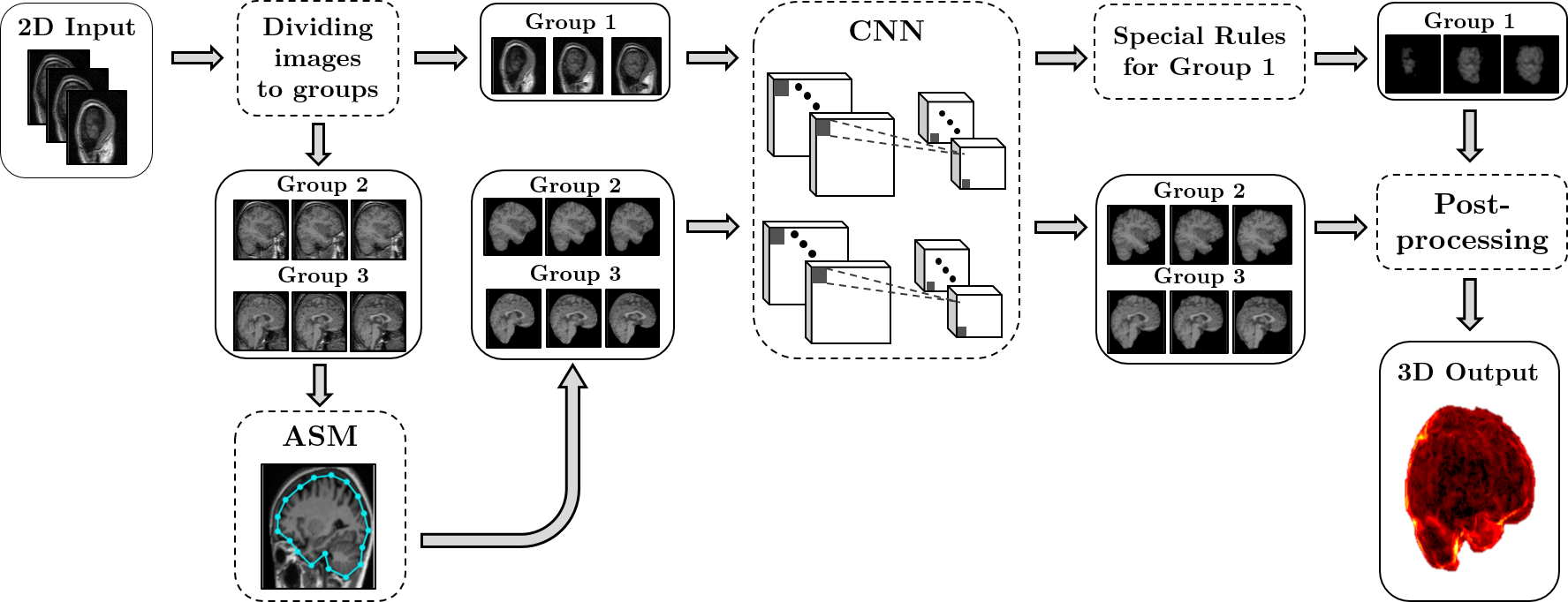}
	\caption{An overview of our proposed method with 2D input and stacking predicted results to obtain a 3D volume. Firstly, images with similar shape and appearance are divided into three distinct groups, namely Group 1, Group 2, and Group 3. Secondly, to obtain initial masks for all groups, a modified Active Shape Model (ASM) is then applied for Group 2 and Group 3, while a Convolutional Neural Network (CNN) is utilized in Group 1 due to its complex shape with disconnected regions. Thirdly, over-smooth results in Group 2 and 3 are corrected by the CNN (same architecture used in Group 1) for pixels locating near the object's boundary. With Group 1, prior medical knowledge from brain positions and obtained masks in Group 2, 3 are integrated by special rules to eliminate noisy regions, e.g., skull areas. Finally, all results are post-processed by a conditional random field to alleviate fatal errors by pixel-wise approaches.}
	\label{fig:flow_chart}
\end{figure}
\section{Methodology}
In neuroimaging, one can use a three-plane coordinate system to describe a human body's standard anatomical position. These imaging planes are the transverse plane, the sagittal plane, and the coronal plane. In ASMCNN, we choose to process 2D slices in the sagittal plane as human brains are nearly symmetrical concerning the mid-sagittal plane. It allows us to predict the general shapes of brain regions with better accuracy. ASMCNN comprises of three main stages: \KH{(1)} dividing images into groups, \KH{(2)} applying ASM to detect initial brain boundaries for images in specific groups, and \KH{(3)} constructing CNN for precise segmentation, along with performing different post-processing methods for different groups to refine segmentation contours. \KH{The flowchart of our system is given in Figures \ref{fig:flow_chart}}. Three main stages of the proposed method are presented in the following subsections.

\subsection{Sagittal-slice classification}

\begin{figure}[!h]
	\centering
	\includegraphics[width=0.5\textwidth]{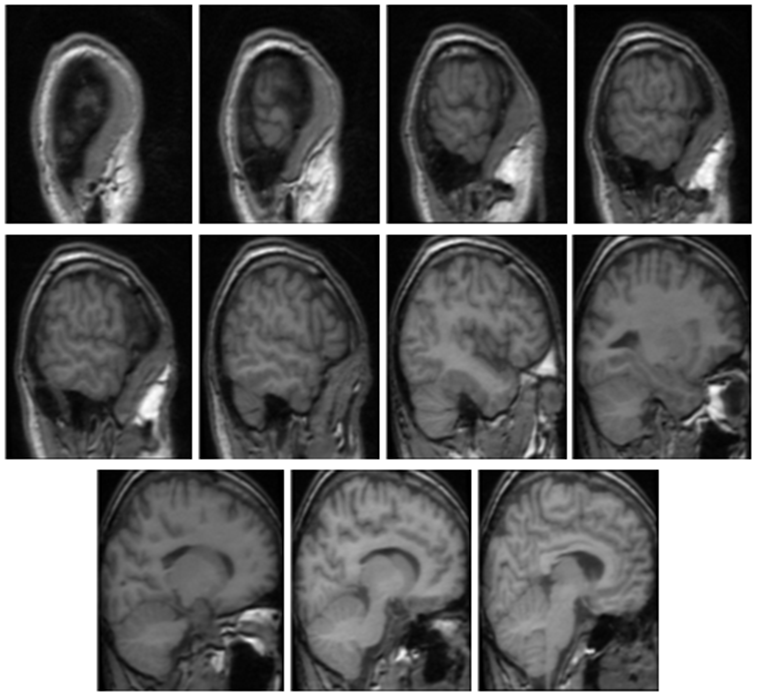}
	\caption{Typical appearances of each group. Top row: group I; middle row: group II; bottom row: group III.}
	\label{fig:group_all}
\end{figure}

\begin{figure}[!h]
	\centering
	\includegraphics[width=0.55\textwidth]{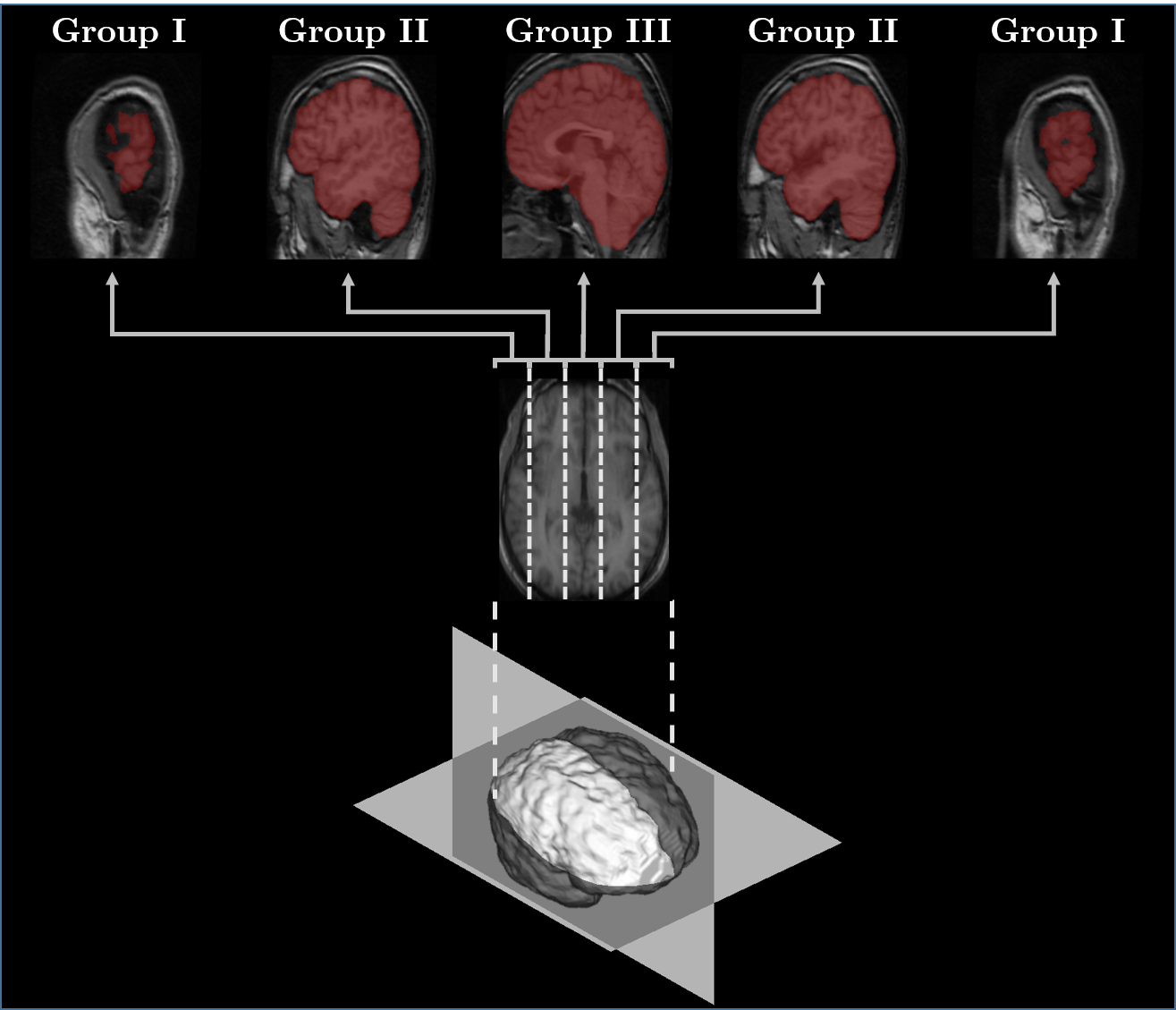}
	\caption{The illustration of the order of three groups in the sagittal plane.}
	\label{fig:three_axis}
\end{figure}

In the first stage, we divide sagittal slices into several groups by initial analysis of brain regions' shapes. We categorize slices into three groups based on their general shape, entitled Group I, II, and III. In Figure \ref{fig:group_all}, we illustrate a few slices in each groups. It is clear that the brain region in Group I is quite small compared to the image size. Meanwhile, the brain areas in Group II and Group III are relatively large, and the brain regions in Group III extend out at the lower-left corner of images. For taking advantage of the symmetry of the human brain across the sagittal plane, we assign the corresponding group of each slice by the following order: Group I, Group II, Group III, Group II, and Group I. Figure \ref{fig:three_axis} depicts how three groups distribute throughout the sagittal axis. 

\begin{figure}[!h]
	\centering
	\includegraphics[width=0.6\textwidth]{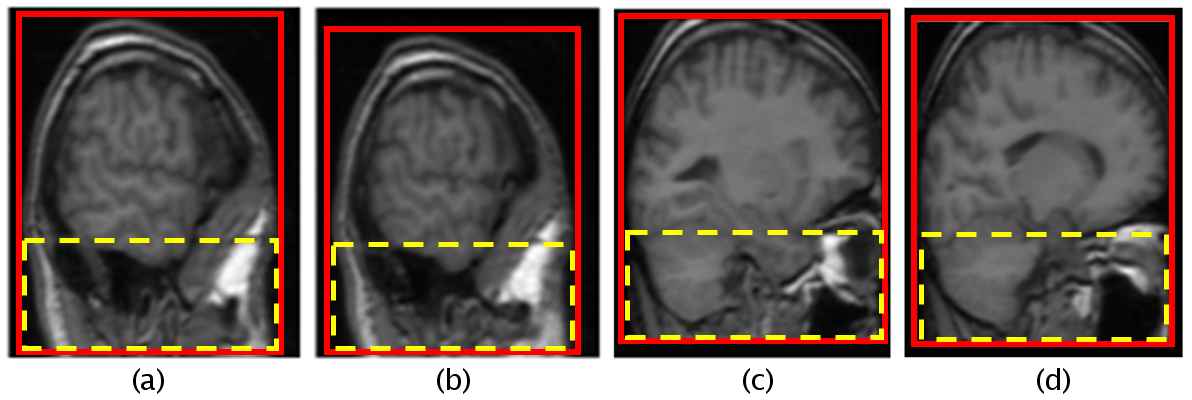}
	\caption{Sub-rectangles and differences between brain shapes in different groups.}
	\label{fig:group_hog}
\end{figure}

We construct a grouping method by using histograms of oriented gradients (HOG) \citep{dalal2005} extracted from a sub-rectangle to train a soft-margin Support Vector Machine (SVM) classifier \cite{chen2004support} (Algorithm \ref{alg:Training}). The sub-rectangle in Algorithm \ref{alg:Training} is illustrated as yellow rectangles in Figure \ref{fig:group_hog}, which are 30\% area at the bottom of the red bounding boxes, indicating the signature differences between brain shapes. Figures \ref{fig:group_hog}a and \ref{fig:group_hog}b are two adjacent slices from Group II and Group I, respectively. As shown in these figures, the brain in Figure \ref{fig:group_hog}a has a small tail at the lower left part, while the brain in Figure \ref{fig:group_hog}b does not. This example shows that brain shapes can be significantly different even though the slices are adjacent. Similarly, Figures \ref{fig:group_hog}c and \ref{fig:group_hog}d are two adjacent slices from Group II and Group III. Here, the brain shape in Group II is more convex, while the one in Group III is concave at the lower right part. We utilize these differences to construct a robust classification method using SVM. Algorithm \ref{alg:Testing} can be used to classify sagittal slices of MRI volumes into different groups by using the SVM model trained by Algorithm \ref{alg:Training}. The notations of Algorithms \ref{alg:Training} and \ref{alg:Testing} are depicted in Table \ref{table:t1}.

After grouping images, we perform various procedures for each group. For Group I, as the brain shape is complex and the brain regions are smaller than those in other groups, it is challenging to propose a general model for brain shapes. Therefore, we use CNN to process all pixels in the images, directly predict the brain regions. We utilize special rules and the Gaussian process to refine the results from the CNN further after predicting with CNN. For Group II and III, we first apply ASM to produce the initial brain boundaries. Then pixels locating around the boundaries are fed to a CNN for precise segmentation. Finally, we perform post-processing using CRF for all groups, producing high-quality segmentation results. In the next section, we present the details of ASM for estimating initial brain contours for Group II and Group III.

\subsection{Active shape model with optimal features}
We apply ASM to slices in Group II and III, producing rough segmentation of brain regions in this stage. For enhancing the modeling capacity of gray-level variations around the border, we use a modified version of ASM called Active Shape Model with optimal features (ASM-OF) \cite{Cootes1995}. It is worth noting that optimal displacements for landmarks in the original ASM algorithm can be replaced by using a nonlinear k-nearest neighbor classifier (k-NN, \cite{Altman1992}) instead of using the linear Mahalanobis distance \cite{van2002active}. We denote all necessary annotations used in the ASM-OF algorithm in Table \ref{table:t2}. This stage comprises three subroutines, and we illustrate each subroutine in the following paragraphs.
\begin{center}
\begin{table}[!h]  
    \begin{tabular}{|c|c|} 
    	\hline
	\bf{Notation} & \bf{Remark} \\
	    	\hline
    $s$ & \makecell{the number of training images.}\\
	\hline
     $n$ & \makecell{the number of landmark points used to represent each image.}\\ 
	\hline
     $n_{s}$ &  \makecell{the number of new positions evaluated during searching.} \\
     	\hline
     $k$ & \makecell{the number of points in profile located either inside and \\outside of the landmark point.}\\
        \hline 
     $l_{max}$ & \makecell{the number of resolution levels for appearance model.}\\
	\hline 
	$n_{grid}$ & \makecell{the size of $n_{grid} \times n_{grid}$ of points, which are sampled \\in training process.}\\
	\hline
	$n_{max}$ &  \makecell{the number of iterations per resolution level.}\\
        \hline   
        $k_{NN}$ &  \makecell{the number of neighbors used in k-NN classifier \\during searching and feature selection.}\\
        \hline   
        $f_{v}$ &  \makecell{a chosen proportion of  the shape variance in \\the training dataset.}\\
        \hline    
     \end{tabular}
    \caption{List of notations used in Active Shape Model with optimal features.}
    \label{table:t2}
\end{table}
\end{center}
\subsubsection{Our proposed shape model}
Normally, a brain image with $s$ sagittal images can be represented by a set of vectors $X_{i}$ by stacking $n$ landmarks $((x_{1},y_{1}),\ldots(x_{n},y_{n}))$ as:
\begin{equation}
X_{i}=(x^i_{1},y^i_{1},\ldots,x^i_{n},y^i_{n})^{T}\ \textrm{where }i\in[1,s].
\end{equation}
One can use the principal component analysis (PCA) method to calculate the mean shape $\overline{X}_{s}$, the corresponding covariance matrix $C$, the first $t$ largest eigenvalues $\lambda_m$ and eigenvectors $\phi_{m}\,(m=1,\ldots,t)$ of $C$. 
After that, a brain shape in the training data set can be approximated by:
\begin{equation}
X_{s}\approx\overline{X}_{s}+\mathbf{\Phi_{s}}\mathbf{b_{s}},\label{eq:update_s}
\end{equation}
where $\mathbf{\Phi_{s}}=(\phi_{1}\ldots\phi_{t})$ is the first $t$ eigenvectors and $\mathbf{b_{s}}=\Phi_{s}^{T}(X_{i}-\overline{X}_{s})\label{eq:bs}$ is a set of shape parameters, $\mathbf{b_{s}}$ can be bounded by 
\begin{equation}
-q\sqrt{\lambda_{i}}\leq\mathbf{b_{s}\leq}q\sqrt{\lambda_{i}}\ (i=1,...,t),\label{eq:limit}
\end{equation}
where $q\in [2,3]$ \cite{van2002active}. The value of the parameter $t$  can be chosen so as to explain a certain proportion $f_{v}$ of the variance in the training shapes, which  usually has a range from $90\%$ to $99.5\%$. For a given $f_{v}$, one can experimentally choose the smallest non-negative integer $t$ such that
\begin{equation}
\sum_{i=1}^{t}\lambda_{i}\geq f_{v}\sum_{i=1}^{2n}\lambda_{i}.
\end{equation}
\subsubsection{A gray-level appearance model with optimal features}
We use a gray-level appearance model to describe a typical image structure around each landmark by applying a k-NN classifier from pixel profiles. These are sampled by using the linear interpolation method around each landmark, which is perpendicular to the contour. From each training image, for each landmark, we define a square grid of $n_{grid}\times n_{grid}$ points in which $n_{grid}$ is chosen as an odd integer, and the landmark point is at the center of the grid. In experiments, $n_{grid}$ is selected as 5; hence, a feature vector has 60 elements, which are calculated at 25 points. The output of each feature vector is a binary value, which is 1 in the case it is inside the object or 0 for other cases. The k-NN classifier with the corresponding weight to each vote is $\exp(-d^{2})$, where $d$ is the Euclidean distance to each neighbor in the feature space. Mann–Whitney algorithm \citep{Mann1947} is used to select the best feature.

Given an input image, each position along the profile is calculated, yielding 60 feature images and processed again to create optimal features. These features are then fed into the k-NN classifier to determine the probability of being inside the object for this pixel. Then, we determine the point $g_{i}$ in the set of points of the profile $g$ such that the objective function $f(g)$ is minimized as
\begin{equation}
f(g)=\sum_{i=-k}^{-1}g_{i}+\sum_{i=0}^{+k}(1-g_{i})\label{eq:minimize}
\end{equation}
where $g_{i}\in [-k,k]$ is oriented from the outside to the inside of the object.
\subsubsection{Model evolution}
After the shape model and the k-NN classifier are constructed from the training dataset, the ASM-OF can be applied to segment objects as follows.
\begin{itemize}
\item Step 1. Initialize the shape model with $\overline{X}_{s}$.
\item Step $2$. For each landmark, put it at $2n_{s}+1$ new locations and evaluate the equation (\ref{eq:minimize}) with the k-NN classifier to find and move it to a new position denoted is $X_{New}$.
\item Step $3$. Fit the shape model by calculating $\mathbf{b_{sNew}}$ using (\ref{eq:bs}) as (\ref{eq:update}), and limit values of $\mathbf{b}_{\mathbf{sNew}}$using (\ref{eq:limit}):
\end{itemize}
\begin{equation}
b_{sNew}=\mathbf{\Phi_{s}^{T}}(X_{New}-\overline{X}_{s})\label{eq:update}
\end{equation}
where $X_{New}=(X_{1New},X_{2New},\ldots,X_{nNew})$.
\begin{itemize}
\item Step $4$. Update the shape landmarks using (\ref{eq:update_s}) as
\end{itemize}
\[
X_{sNew}\approx\overline{X}_{s}+\mathbf{\Phi_{s}}\mathbf{b_{sNew}}
\]
\begin{itemize}
\item Step $5$. Iterate steps $2$ and $4$ up to a predefined $n_{max}$
times.\end{itemize}

Since ASM-OF can only capture general shapes of the brain region, derived boundaries are very smooth. Therefore, in the next stage, we utilize a CNN for exact segmentation of results from ASM-OF to accurately localize the boundary of the brain region.
\subsection{Convolutional neural networks}
\label{sec:cnn}
In this section, we present the structures of the CNNs. For Group I, one CNN focuses on all pixels in the brain image and directly identifies pixels belonging to the brain regions. For Group II and III, we construct another CNN which focuses on pixels located around the boundary of the preliminary brain mask estimated from the previous stage instead of using all pixels. Additionally, we also exploit the global spatial information \citep{Liu2014} and combine it with the feature maps learned from CNN as an input vector for multi-layer perceptrons. We describe the architecture of the proposed CNN in the following sections.
\subsubsection{Input and output space}
\begin{figure}[!h]
	\centering
	\includegraphics[width=0.2\textwidth, scale = 0.3]{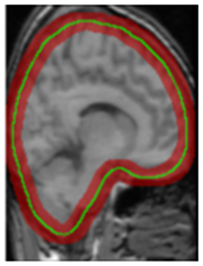}
	\caption{Preliminary brain boundary produced by ASM (green) and surrounding pixels that are classified by CNN (red).}
	\label{fig:extract_f}
\end{figure}

Assume that $N$ is the number of slices of an MRI volume, $I_{1}=\{I_{11},...,I_{1p}\}$, $I_{2}=\{I_{21},...,I_{2q}\},\, I_{3}=\{I_{31},...,I_{3r}\}$ are the sets of images in Group I, II, and III respectively, where $p+q+r=N$, $M_{2}=\{M_{21},...,M_{2q}\},\, M_{3}=\{M_{31},...,M_{3r}\}$ are the sets of brain masks produced by ASM for Group II and III. For each image in Group II and III, positions of pixels around the boundary of $M_{i}(i\in\{2,3\})$ with the distance around 5 pixels are extracted by using the algorithm \ref{alg:MergeSlice} (Figure \ref{fig:extract_f}). These positions are then extracted the information based on $I_{j}\,(j\in\{2,3\}).$ With images in $I_{1},$ all pixels within the rectangle, which contains the skull, are used as features. We aim to employ CNNs to predict each pixel belonging to a brain or non-brain region. 

To this end, we extract two following features representing each pixel. First, we consider local features, which are three adjacent image slices of the size $11 \times 11$ centered around each pixel. Next, we calculate global features by combining the position $(x,y)$ of each pixel and the index $z$ of the corresponding image where pixel belongs to $(1\leq z\leq N)$. Figure \ref{fig:struct_deep} illustrates detailedly the feature extraction process and how to combine two features in the network.

\subsubsection{A proposed CNN architecture}
\begin{figure}
	\includegraphics[width=\textwidth]{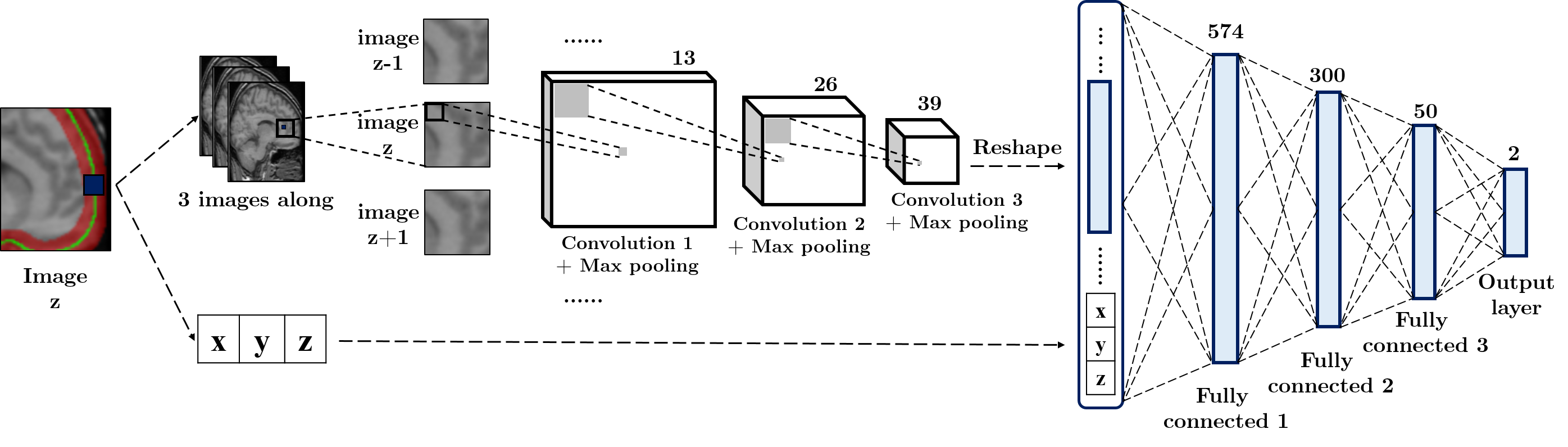}
	\caption{Feature extraction step and the proposed deep neural network structure.}
	\label{fig:struct_deep}
\end{figure}

Figure \ref{fig:struct_deep} describes our network structure. It includes three convolutional layers followed by four fully connected layers, including the final two-node layers for the output. The size of the convolutional layers is $3\times3$, and the max-pooling layers use $2\times2$ kernel. We choose the depths of the first, the second, and the third convolutional layers as $13$, $26$, and $39$, respectively. The ReLU activation function ($f(x)=\max(0,x)$) is applied to outputs of the convolutional layer.

After that, three vectors from three slices of images obtained from convolutional layers can be concatenated with the normalized coordinates to form a new vector. This vector is later fed into the fully connected layers. The depths of four fully connected layers are $574$, $300$, $50$, and $2$, consecutively. The final layer has the size of two, indicating the probability of the input pixel belonging to the brain or non-brain region. 
\subsubsection{Training and Optimization}
The aim of CNN is to minimize the $L_2$ penalized cost function:
\begin{equation}
L(w)=\frac{1}{n}\sum_{i=1}^{n}l(z,f(x,w))+\frac{\eta}{2}||w||^{2},\label{eq:Fun_cost}
\end{equation}
for a labeled training set $(x_{i},z_{i}),$ $i\in(1,n)$ and weights $(w_{1},\ldots w_{L})$ including convolutional kernel weights and bias weights with a loss function $l$ . The loss function $l$ can be defined as: 
\begin{equation}
l(z,f(x,w))=-\frac{1}{N_1}\sum_{n=1}^{N_1}\sum_{m=1}^{M_1}B_{m,n}\log(p_{m,n})
\end{equation}
where $N_1$ is the number of training images in the batch, $M_1$ is the number of classes, and $p_{m,n}$ is the probability of the $n$-th sample being classified into the $m$-th class. If the $n$-th sample is classified into the $m$-th class, $B_{m,n}$ equals to $1$, otherwise $B_{m,n}$ is 0. 
We fix $\eta = 0.005$ and learn the weights by using Adam optimizer. All weights in each layer at the convolutional layer are initialized from a normal distribution of $N(0,0.1)$ while weights in multi-layer perceptrons are created from a normal
distribution of $N(0,\frac{1}{n_{hidden}})$   
where $n_{hidden}$ is the number of hidden layers of the previous layer \cite{he2015delving}. The CNN is trained by using mini-batch stochastic gradient descent with a batch size of $128$. For preventing over-fitting, we use a dropout layer after each layer in the fully connected layers with
the rate $50\%$.
\subsection{Special Rules for Group I} \label{sec:group-i}
Brain regions in Group I are small and noisy, which might be cases that CNNs produce two regions in the segmented results, as demonstrated in Figure \ref{fig:group01final}. However, it is difficult to determine which area is the brain region automatically. To overcome this, we utilize Gaussian processes \cite{rasmussen2006cki}. Gaussian processes are used to learn the change in positions of the brain center points throughout sagittal slices. Based on this change, we can predict the center of cerebral images in Group I. Then, the area closest to the center of the gravity is selected for execution.

\begin{figure}[!h]
	\centering
	\includegraphics[scale = .4]{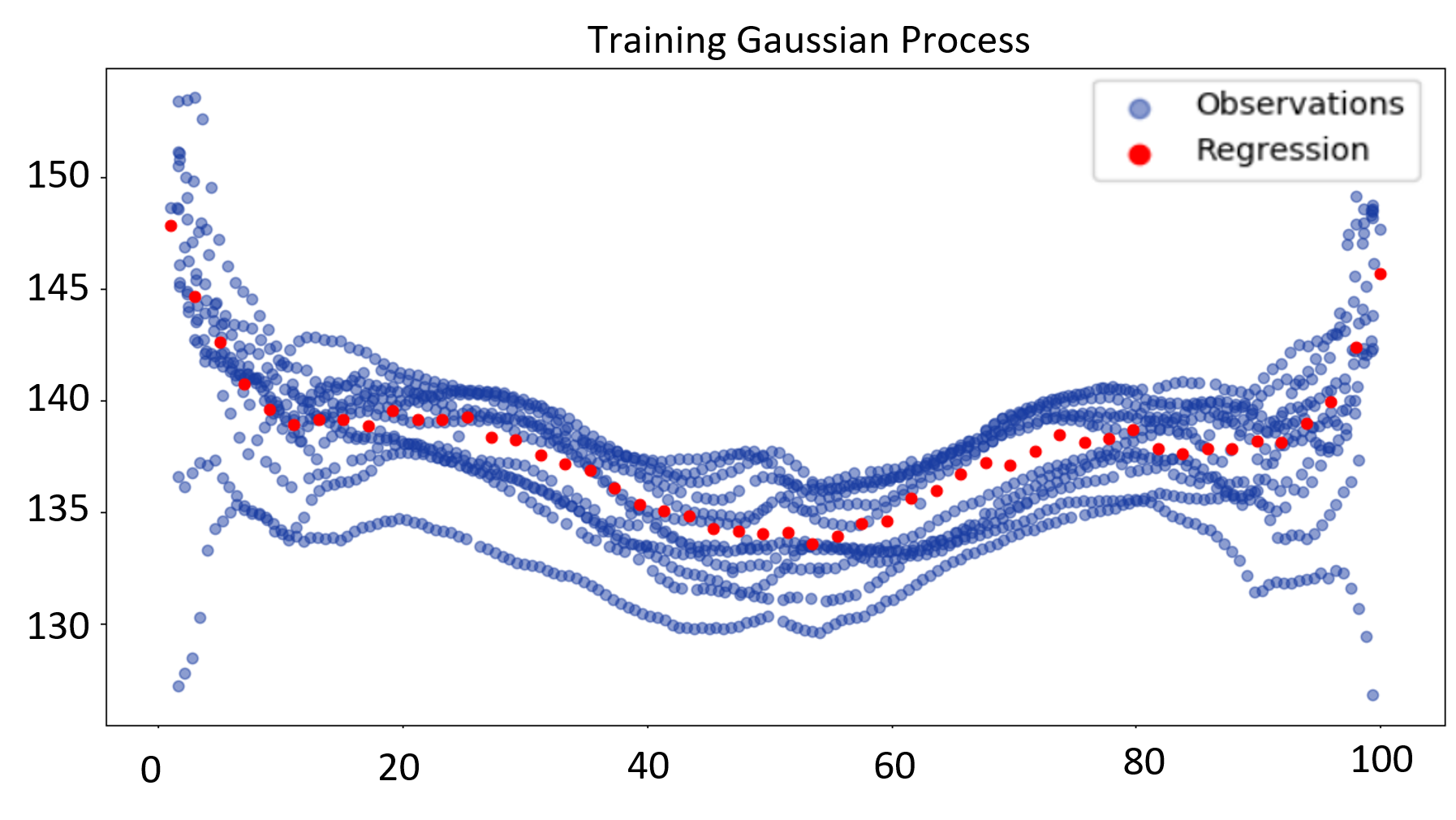}	
	\caption{Illustration for training process of Gaussian processes in the IBSR dataset.}
	\label{fig:gauss_train}
\end{figure}

Figure \ref{fig:gauss_train} illustrates training results with the IBSR dataset \cite{IBSR} by Gaussian processes. The blue lines illustrate the change of the center position of cerebral ventricles for each subject from the starting position to the ending position; here, 12 blue lines indicate 12 subjects. It is worth noting that the positions of the starting slice and the ending slice in MRI volumes are not consistent. Therefore, we normalize the starting and ending slice indexes to $[1, 100]$ as visualized on the x-axis (Figure \ref{fig:gauss_predict}). The y-axis represents the norm-2 of the center position of the brain images from the training data. The red line illustrates the result of Gaussian processes at 50 different normalized slice indexes.

Figure \ref{fig:gauss_predict} depicts the predicted results by Gaussian processes. The red line illustrates the Gaussian processes' initial prediction for a new subject. After adjusting this prediction, we obtain the blue line, which is the final result of the Gaussian processes. The purple line is the ground truth. The adjustment for predictions from the Gaussian processes is based on two factors:

\begin{figure}[!h]
	\centering
	\includegraphics[scale=.4]{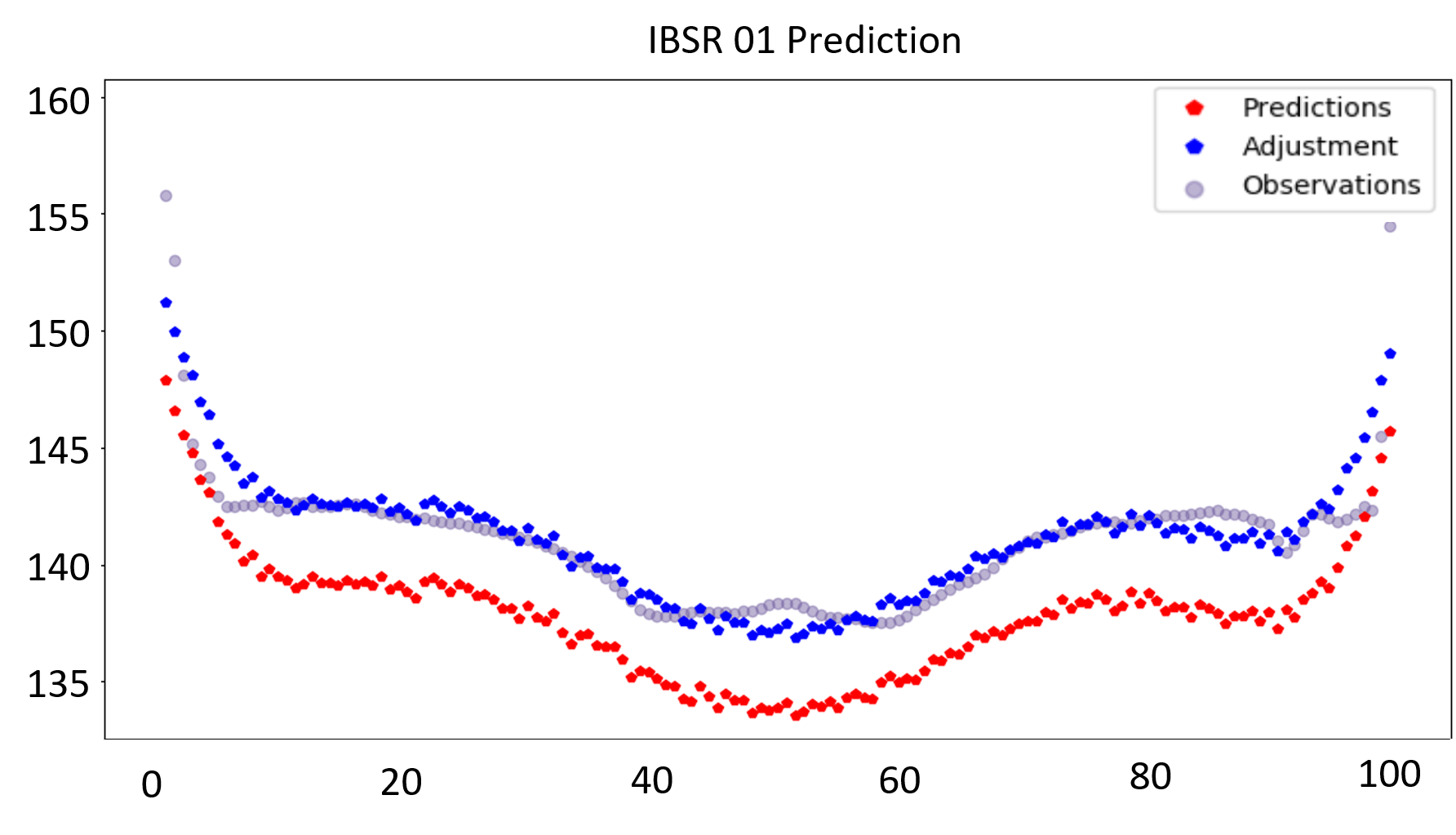}	
	\caption{Predicted results of Gaussian processes in the IBSR dataset.}
	\label{fig:gauss_predict}
\end{figure}

\begin{itemize}
\item  By various image acquisition conditions, an image of a subject can be shifted but usually, the changing rate of the image center along each slice may remain unchanged. Hence, it is possible to correctly predict the center of the brain region by translating the initial prediction with a value $\alpha$.
\item For each subject, one can firstly combine ASM-OF, CNN, and CRF to detect two regions II and III. Next, one can consider centers of all images in both two regions, compare the deviation among predicted results by Gaussian processes, and compute the mean value as a desired value of $\alpha$.
\end{itemize}

After doing the translation and obtaining the blue line, for each slice in Group I, we calculate the distance between each connected component and the predicted value of Gaussian processes and select the region with the shortest distance. The algorithm \ref{alg:PIG1} describes  detailedly how images in Group I are processed. The list of its sub-algorithms including  \textit{CheckCenter} (\ref{alg:Check_Center}), \textit{CheckArea} (\ref{alg:Check_Area}),\textit{ConvertRange} (\ref{alg:Convert_Range}), and \textit{CheckDistance} (\ref{alg:Check_Distance}) are presented in Appendices  and \textit{CRF} is presented in the section \ref{sec:Post-processing}.
\subsection{Post-processing}
\label{sec:Post-processing}
In this stage, we apply post-processing techniques for further refining segmented slices from all groups. Because all main procedures are pixel-wise approaches, segmentation maps are not smooth and precise, which can decrease the overall accuracy of the brain extraction process significantly. To overcome this issue, we utilize a post-processing technique based on the conditional random field (CRF) to validate the extracted results from both ASM-OF and CNN and to refine these segmentation maps.

Let $\mathbf{I}$ be an image of size $N_2\times N_2$ and $\mathbf{x}$ be its segmentation map. The Gibbs energy of $\mathbf{x}$ can be given by
\begin{equation}
	E(\mathbf{x}) = \sum_i \varphi_u (x_i) + \sum_{i < j} \varphi_p (x_i,x_j)
\end{equation}
where $x_i$ and $x_j$ are labels assigned to the $i$-th and $j$-th pixels, respectively. Here, the label is 1 when the pixel in the \lq\lq brain\rq\rq\ region and 0 for other cases. The unary potential is given by $\varphi_u (x_i) = -\log P(x_i)$, where $P(x_i)$ is the probability of pixel $i$ getting classified as \lq\lq brain\rq\rq, and is estimated from the output of the main procedure. Denoting $p_i, I_i$ are the position and the grey value features of pixel $i$, 
we use the contrast-sensitive two-kernel pairwise potential $\varphi_p (x_i,x_j)$ as in \citep{krahen2011},
\begin{equation}\label{gaussker}
\small 
\psi_p(x_i,x_j) = \mu(x_i,x_j) \left[ \exp\left( -\frac{|p_i-p_j|^2}{2\theta_\alpha^2}-\frac{|I_i-I_j|^2}{2\theta^2_\beta} \right)+\exp\left(-\frac{|p_i-p_j|^2}{2\theta_\gamma^2}\right)\right],
\end{equation}
where $\mu(x_i,x_j)=[x_i\neq x_j]$ is the Potts model compatibility function that captures the compatibility between different pairs of labels,  parameters $\theta_\alpha$,  $\theta_\gamma$ and  $\theta_\beta$ control the degrees of the nearness and appearance similarity. In our experiments, we set $\theta_\alpha = (8, 11), \, \theta_\beta = (8, 8)$. 

\begin{figure}[!h]
	\centering
	\includegraphics[width=0.68\textwidth]{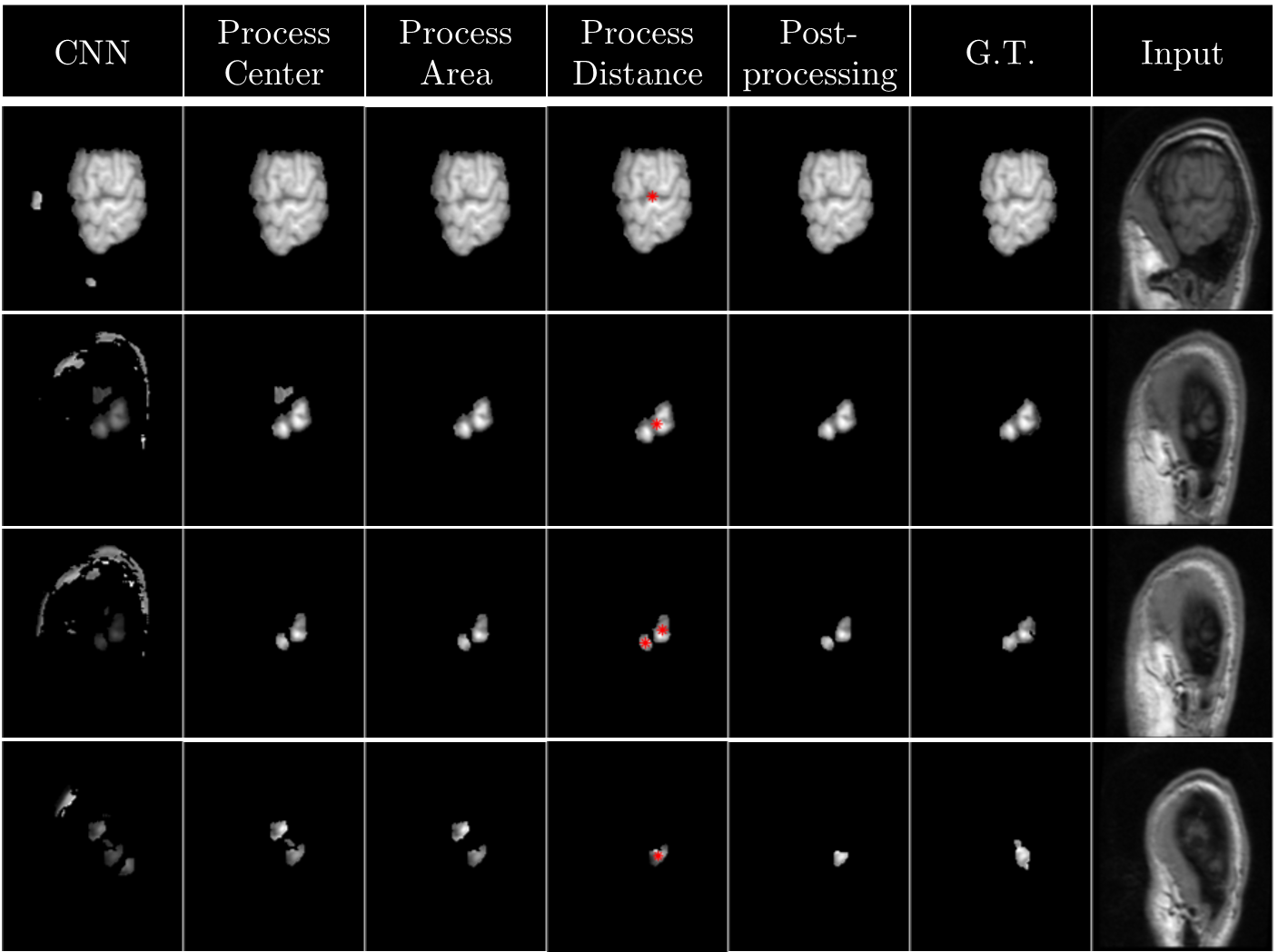}
	\caption{Processing steps for Group I.}
	\label{fig:group01final}
\end{figure}

\begin{figure}
	\centering
	\includegraphics[width=0.7\textwidth]{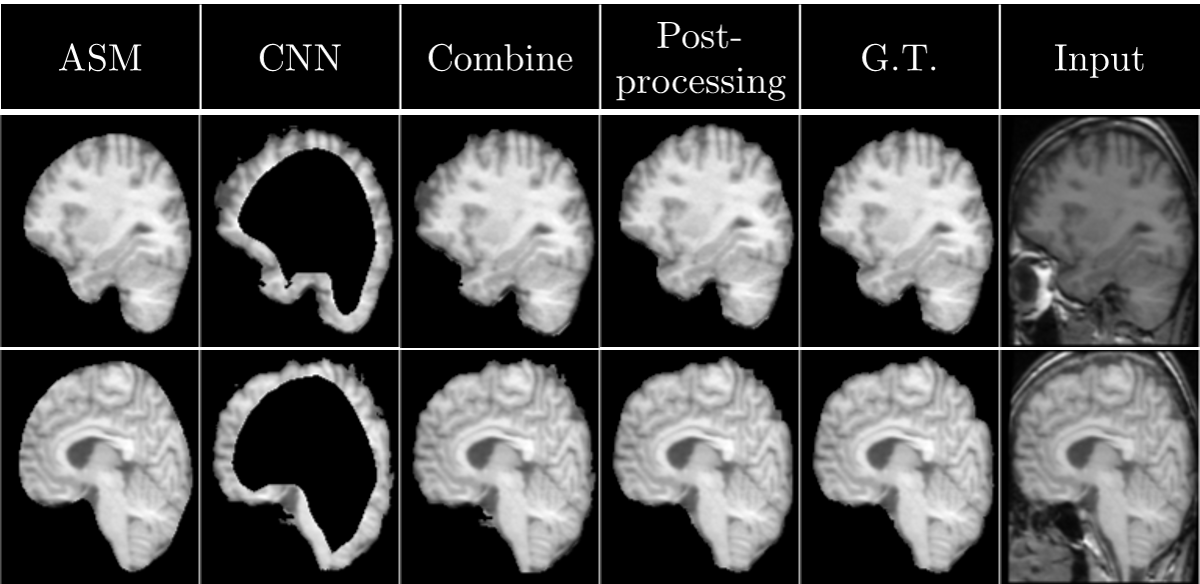}
	\caption{Processing steps for Group II and Group III.}
	\label{fig:group0203final}
\end{figure}

The most probable segmentation map $\mathbf{x}^*$ is calculated by
\begin{equation}
	\mathbf{x}^* = \argmin_{\mathbf{x} \in \mathcal{L}^N} E(\mathbf{x})
\end{equation}
where $E(x)$ is minimized by the mean-field approximation as described in \citep{krahen2011}. The visualization of all stages from CNN to post-processing steps for Group I and Group II, III are illustrated in Figures \ref{fig:group01final} and \ref{fig:group0203final} respectively.

\section{Experiments}
\subsection{ASMCNN versus others}
We compare our proposed approach with seven well-known methods, including Brain Extraction Tool (BET) \citep{Smith2002}, Brain Surface Extractor (BSE) \citep{Shattuck2001}, 3DSkullStrip (3DSS) \citep{Cox1996}, ROBEX \citep{Iglesias2011}, Brain Extraction based on nonlocal Segmentation Technique (BEAST) \citep{Eskildsen2012}, 3D-CNN \citep{kleesiek2016deep}, and U-Net approach \citep{dong2017automatic}. All necessary implementations are optimized to work with T1-weighted data. Based on codes released by Hao Dong et al. \citep{dong2017automatic}, we adopt the proposed network to train end to end with 2D image sequences in the sagittal plane, denoted as 2D-Unet.
 For 3D-CNN by Kleesiek et al., we only extract the final results mentioned in \citep{kleesiek2016deep} for the comparison in the chosen datasets. We evaluate the performance of all these methods by three publicly available datasets: Internet Brain Segmentation Repository (IBSR) \cite{IBSR}, LONI Probabilistic Brain Atlas (LPBA) \cite{LPBA}, and the Open Access Series of Imaging Studies (OASIS) project \cite{OASIS}. We use 6 out of 20 subjects in the IBSR dataset for training and validation in this work, and all 14 remaining subjects are chosen for testing. Similarly, 12/28 is the ratio between the number of training and testing subjects in the LPBA dataset, and 22/55 is the corresponding ratio for the OASIS dataset. 

\begin{figure}[!hbtp]
	\centering	\includegraphics[width=0.85\textwidth]{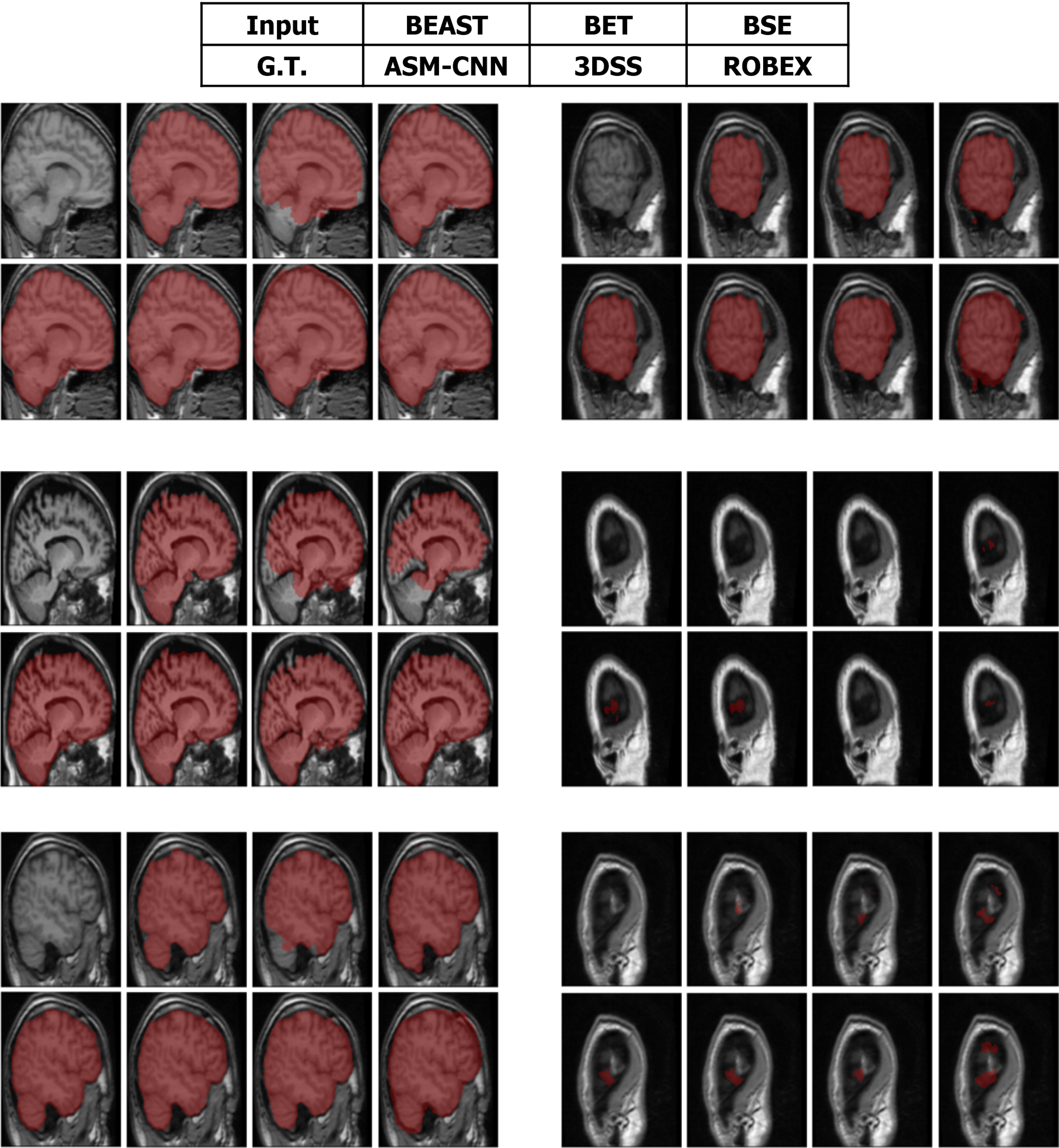}
	\caption{The comparison between ASMCNN with other methods on IBSR dataset.}
	\label{fig:IBSR_Compare}
\end{figure}

\subsection{Settings and metrics}
We implement the proposed system in Python with the TensorFlow library \citep{tensorflow2015-whitepaper} for feature extraction and training deep neural networks. The architecture of deep neural networks is described in Figure \ref{fig:struct_deep}, with the ReLU activation function for all three datasets. To avoid over-fitting, we add dropout layers with a rate of $50\%$ after each layer in a deep neural network. The training phase took approximately 0.8 hours for IBSR, 1.6 hours for LPBA, and 3.2 hours for OASIS to complete. This process has been performed on a workstation equipped with Intel(R) Xeon(R) E5-2698v4 CPU running at 2.20Ghz and an NVIDIA GeForce GTX 1080 GPU. The average time for processing one single MRI volume in the testing phase is approximately 48 seconds. We evaluate all brain extraction methods' performance by comparing segmentation results with the ground truth in each dataset using the Dice coefficient, Jaccard index, Average Hausdorff Distance (AHD), Sensitivity score, and Specificity score \cite{bertels2019optimizing}.

\subsection{Qualitative evaluation}
\begin{figure}[!hbtp]
	\centering
	\includegraphics[width=0.8\textwidth]{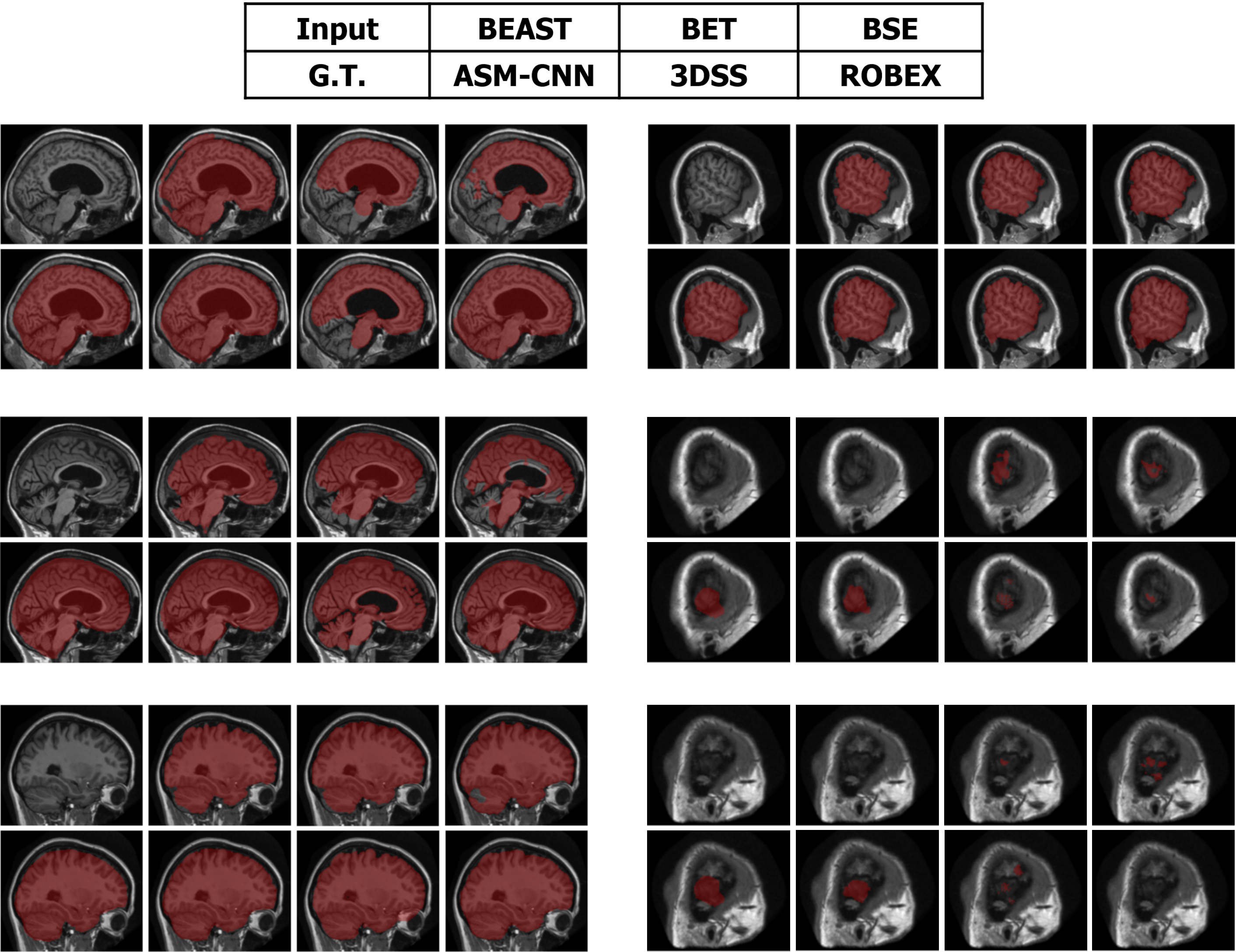}
	\caption{The comparison between ASMCNN with other methods on OASIS dataset.}
	\label{fig:OASIS_Compare}
\end{figure}

\begin{figure}[!hbtp]
	\centering
	\includegraphics[width=0.8\textwidth]{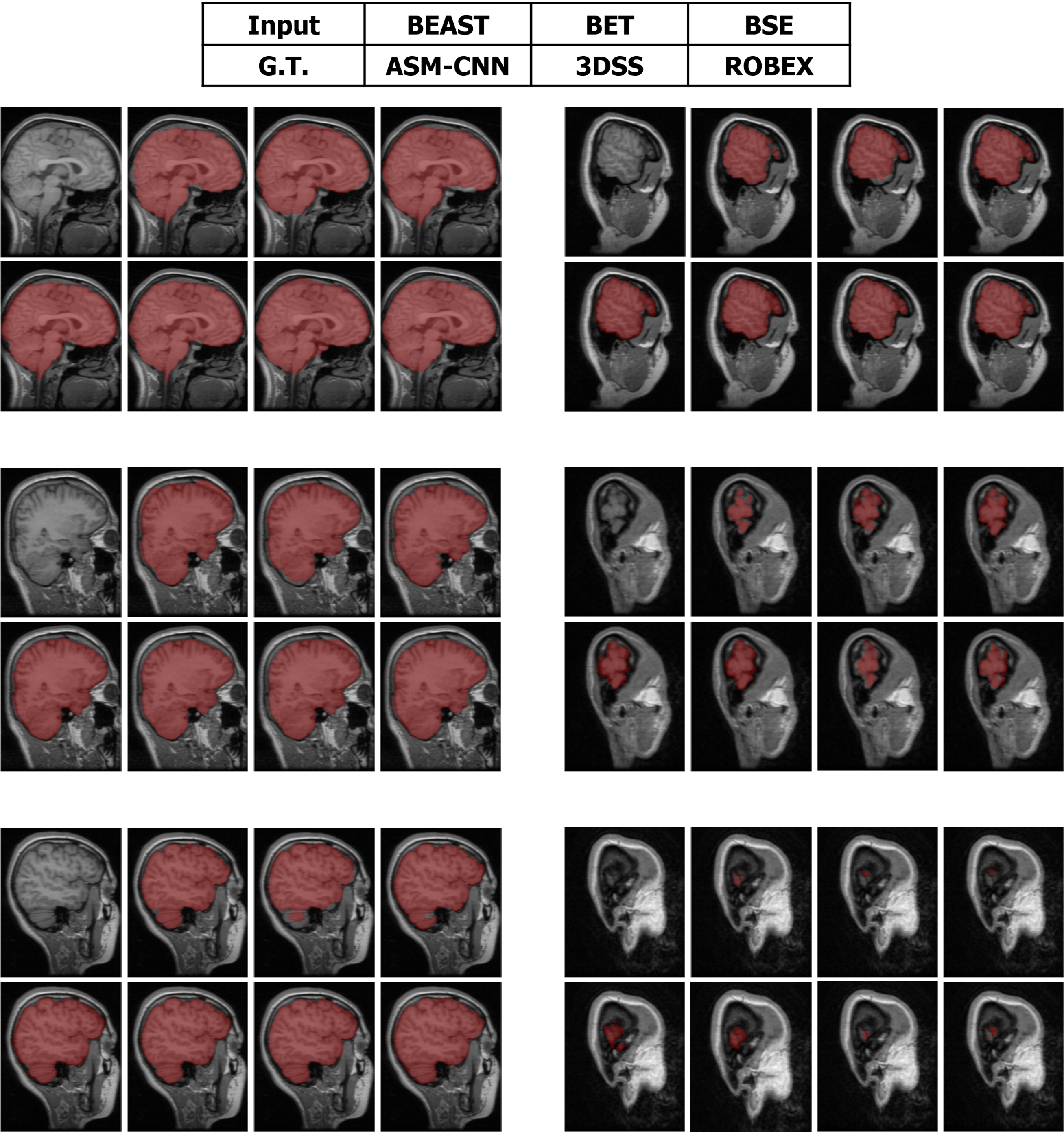}		
	\caption{The comparison between ASMCNN with other methods on LPBA dataset.}
	\label{fig:LPBA_Compare}
\end{figure}

\begin{figure}[!htbp]
	\centering
	\includegraphics[width=0.85\textwidth]{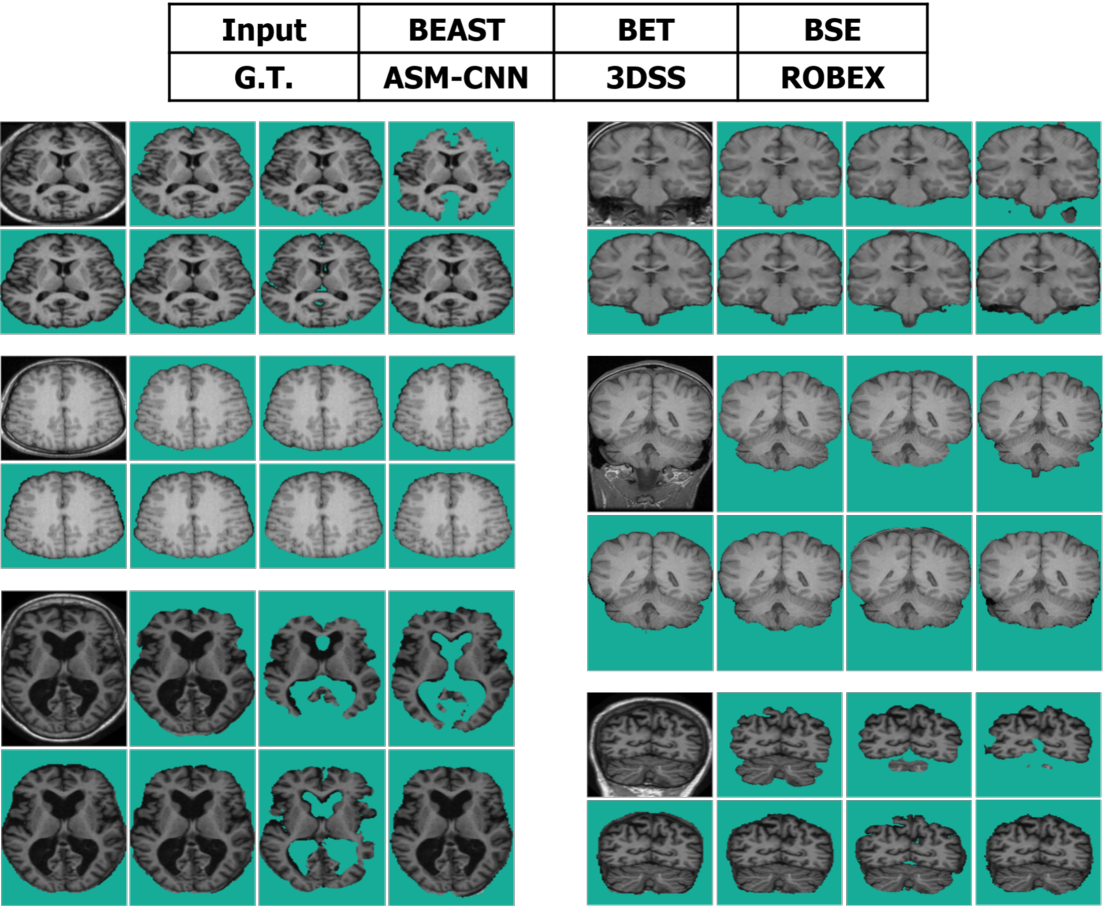}
	\caption{Resulting images on transverse and coronal plane for IBSR, LPBA, OASIS dataset.}
	\label{fig:axial_coronal}
\end{figure}

Segmentation results of ASMCNN and seven other methods in the sagittal plane for the three datasets are illustrated in Figures \ref{fig:IBSR_Compare}, \ref{fig:OASIS_Compare} and \ref{fig:LPBA_Compare} respectively. Each figure includes six typical testing scans from all three groups (two scans for each group). Although ASMCNN works with the sagittal plane, it also produces correct segmentation in two other planes. Figure \ref{fig:axial_coronal} shows the comparison between our approach and other methods on each dataset for these two planes.

ASMCNN can provide extremely accurate segmentation in these three datasets. We achieve the same results as ROBEX with smooth boundaries and keep both gray and dura matter inside extracted brains for most cases, which are usually left out by ROBEX. Although there are minor leakages into the skull by ASMCNN, its occurrence is less than both ROBEX and BEAST, as it only generates a smaller number of over-segmentation results. The critical impact of the method is that it can precisely work for small-size brains in Group I as well, meanwhile other techniques usually fail. As shown in Figures \ref{fig:IBSR_Compare} - \ref{fig:LPBA_Compare}, our results are mostly similar to the ground-truth images especially for tiny-size brains (Figure \ref{fig:IBSR_Compare}). The method does not avoid a few false negatives and false positives in this group on account of the complexity of the brain structure. Nevertheless, ASMCNN can mostly obtain a better performance than others with higher accuracy.

\subsection{Quantitative evaluation}
\begin{figure}
	\centering
	\includegraphics[width=1.0\textwidth, scale=1.5]{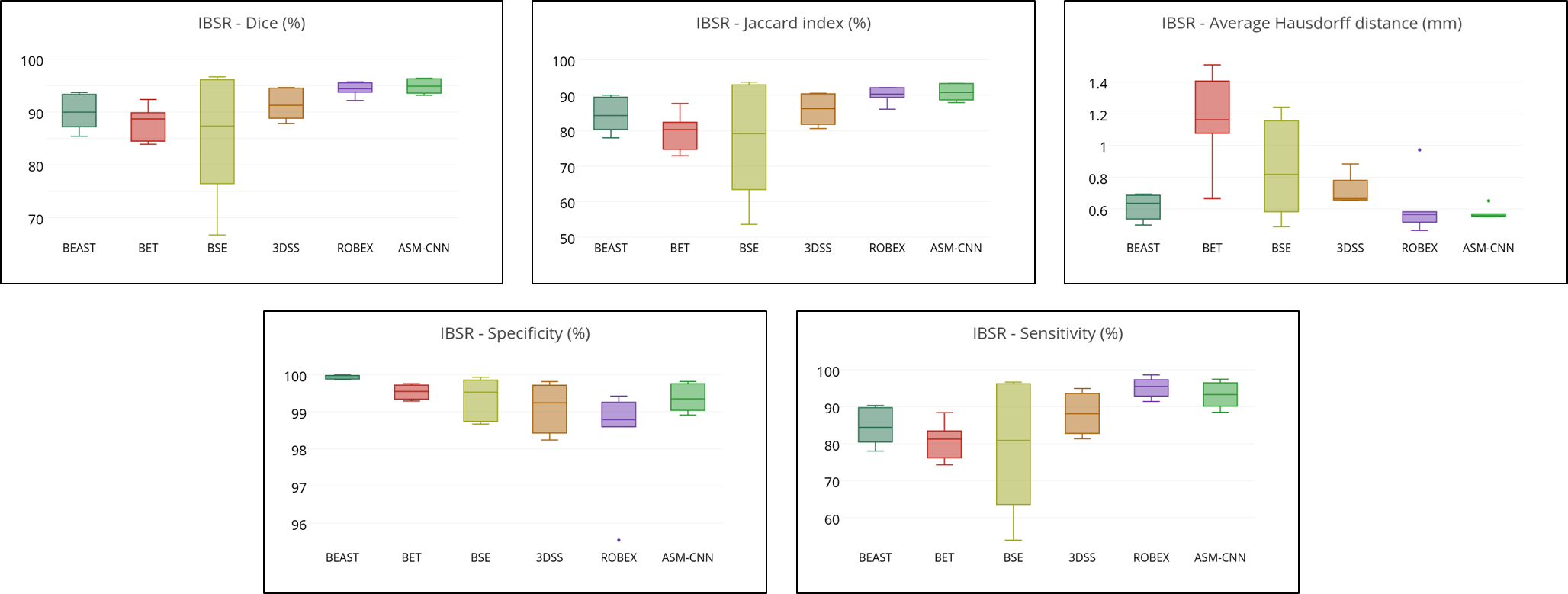}
	\caption{The 2D Box plots of Dice coefficient, Jaccard index, Average Hausdorff distance, Sensitivity and Specificity for IBSR dataset.}
	\label{fig:IBSR_2d_Boxplot}
\end{figure}

\begin{figure}
	\centering
	\includegraphics[width=1.0\textwidth, scale=1.5]{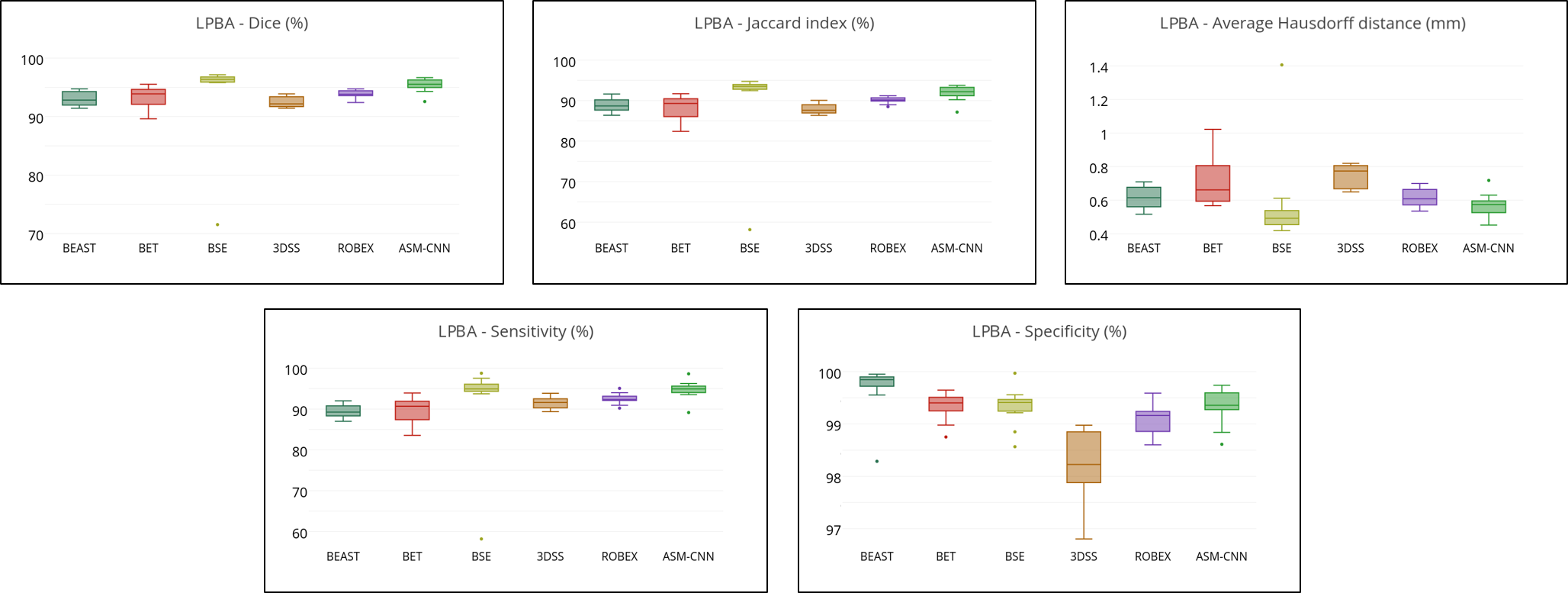}
	\caption{The 2D Box plots of Dice coefficient, Jaccard index, Average Hausdorff distance, Sensitivity and Specificity for LPBA dataset.}
	\label{fig:LPBA_2d_Boxplot}
\end{figure}

\begin{figure}
	\centering
	\includegraphics[width=1.0\textwidth, scale=1.5]{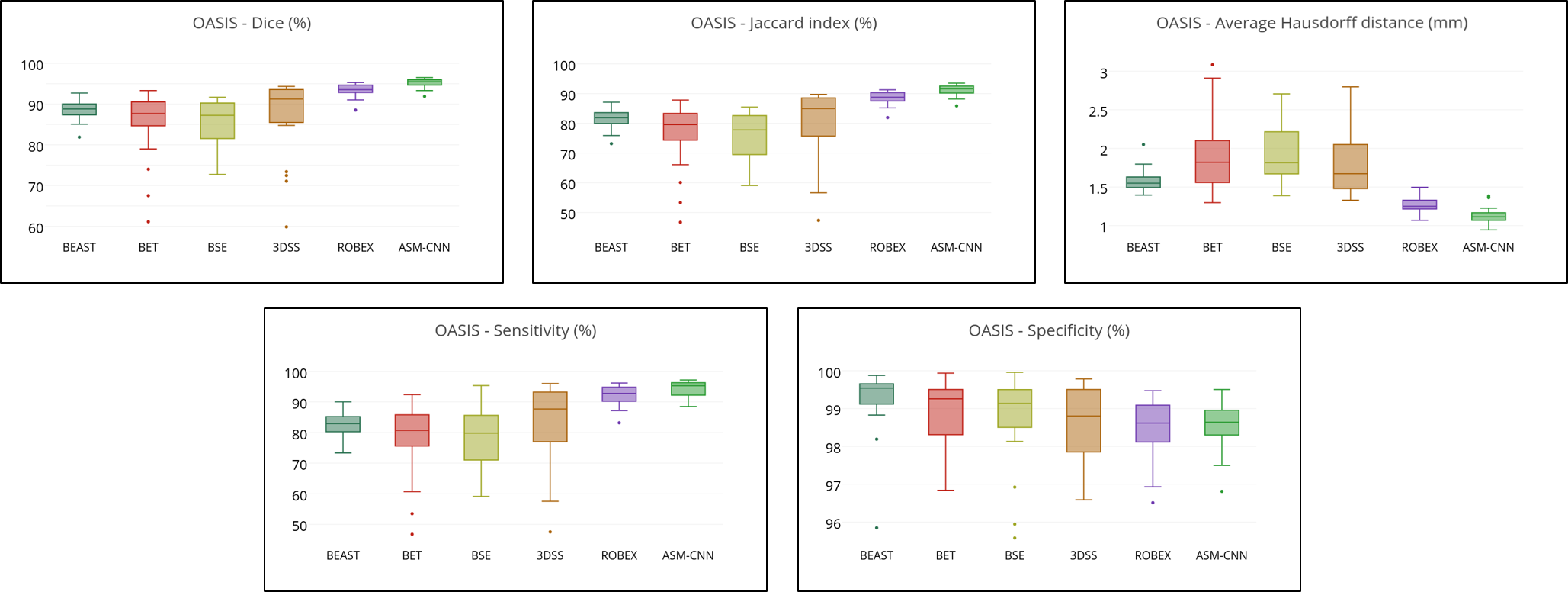}
	\caption{The 2D Box plots of Dice coefficient, Jaccard index, Average Hausdorff distance, Sensitivity and Specificity for OASIS dataset.}
	\label{fig:OASIS_2d_Boxplot}
\end{figure}

\begin{figure}
	\centering
	\includegraphics[width=1.0\textwidth, scale=1.5]{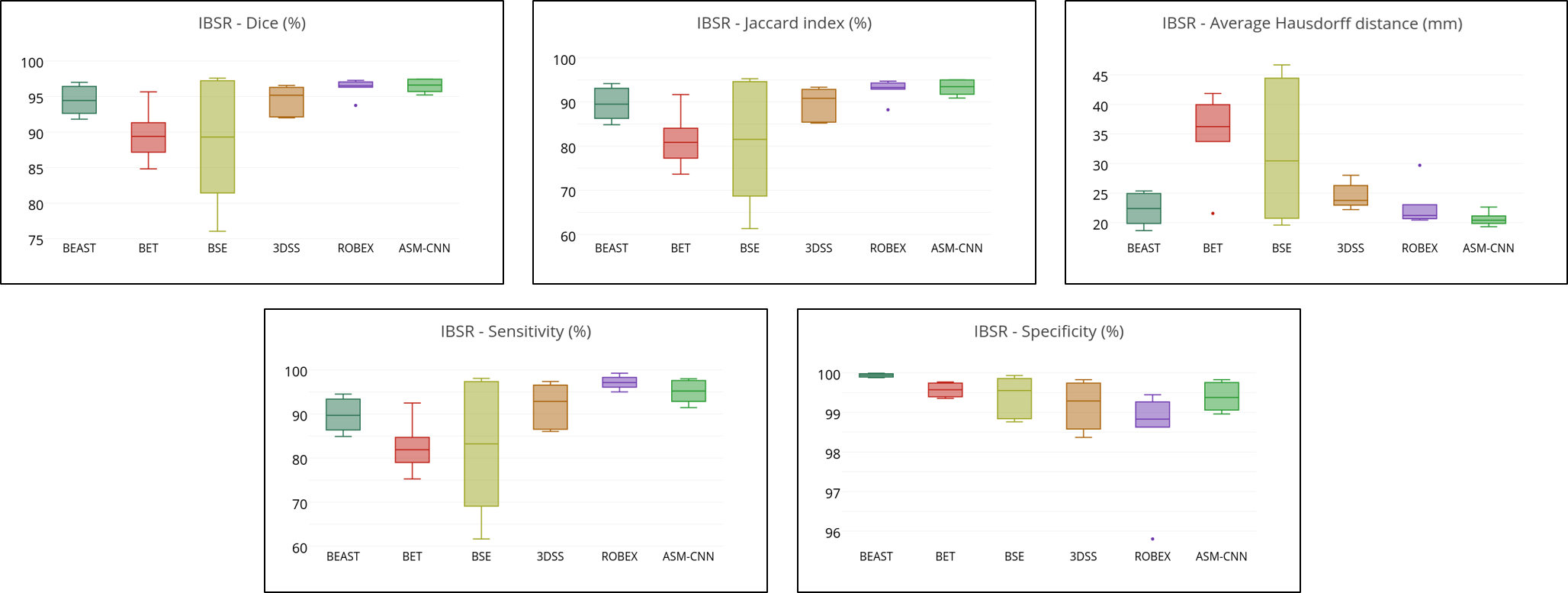}
	\caption{The 3D Box plots of Dice coefficient, Jaccard index, Average Hausdorff distance, Sensitivity and Specificity for IBSR dataset.}
	\label{fig:IBSR_3d_Boxplot}
\end{figure}

\begin{figure}
	\centering
	\includegraphics[width=1.0\textwidth, scale=1.5]{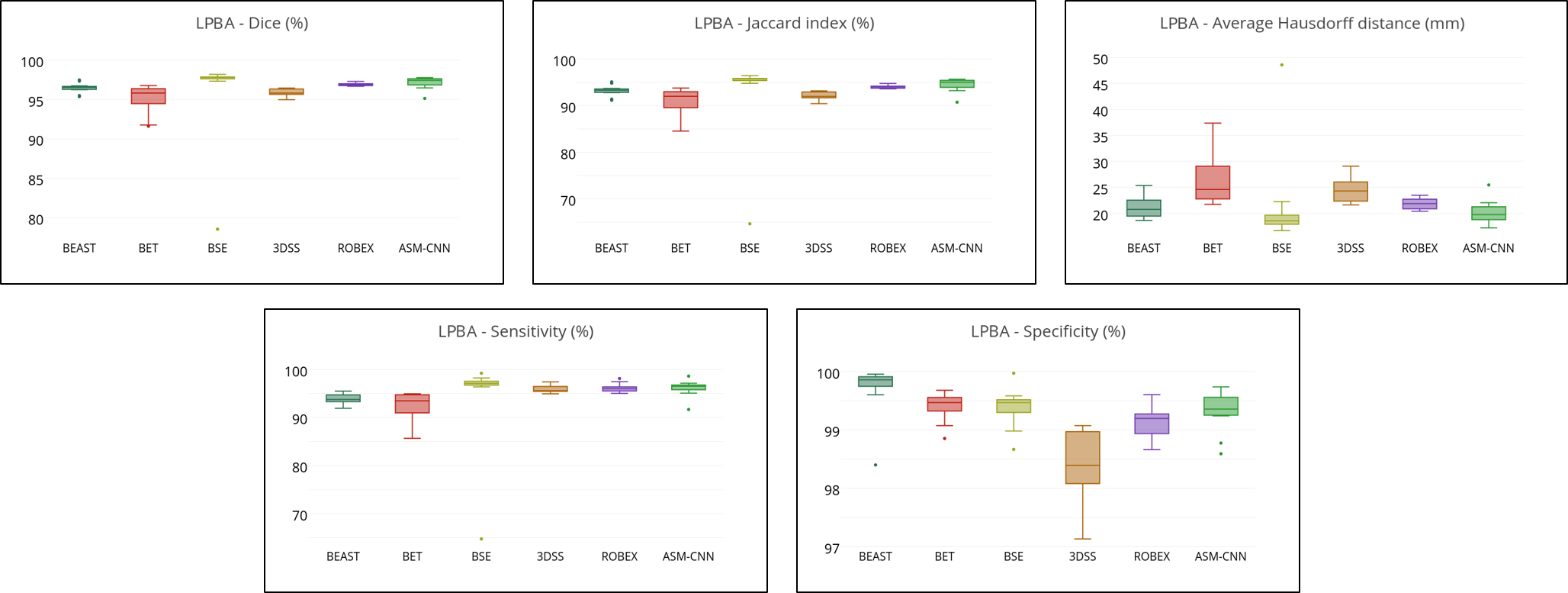}
	\caption{The 3D Box plots of Dice coefficient, Jaccard index, Average Hausdorff distance, Sensitivity and Specificity for LPBA dataset.}
	\label{fig:LPBA_3d_Boxplot}
\end{figure}

\begin{figure}
	\centering
	\includegraphics[width=1.0\textwidth, scale=1.5]{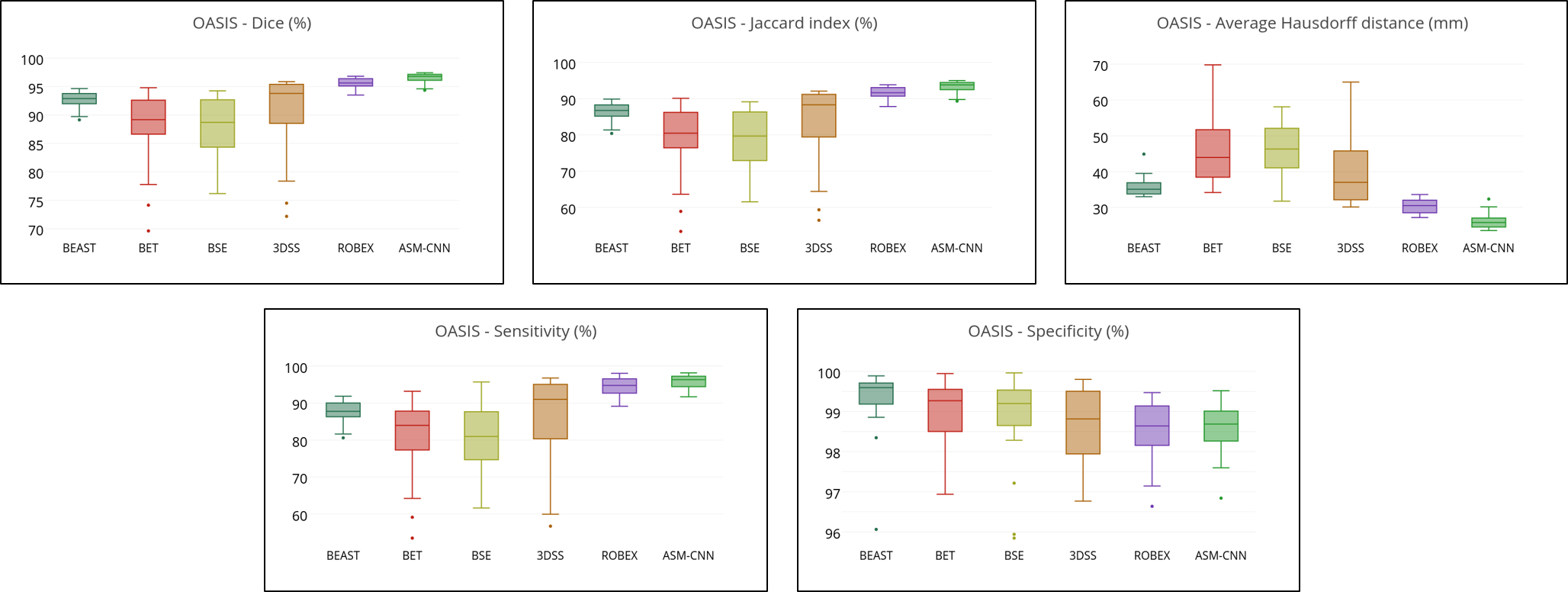}
	\caption{The 3D Box plots of Dice coefficient, Jaccard index, Average Hausdorff distance, Sensitivity and Specificity for OASIS dataset.}
	\label{fig:OASIS_3d_Boxplot}
\end{figure}

Tables \ref{tab:2DIBSR}, \ref{tab:2DLPBA}, and \ref{tab:2DOASIS} display the average value and the standard deviation of all the evaluation metrics for IBSR, LPBA, and OASIS datasets on 2D plane respectively. Similarly, Tables \ref{tab:3DIBSR}, \ref{tab:3DLPBA}, and \ref{tab:3DOASIS} show the accuracy of segmented results on 3D structures. All results by different metrics with the list of evaluation methods on three datasets are depicted by box plots in Figures \ref{fig:IBSR_2d_Boxplot}-\ref{fig:OASIS_3d_Boxplot}. 

\begin{table}[!htbp]
	\centering
	\centerline{
	\scalebox{0.9}{
		\begin{tabular}{llllll}
		\hline 
		\textbf{Method} & \textbf{Dice} & \textbf{Jaccard } & \textbf{Average Hausdorff} & \textbf{Sensitivity} & \textbf{Specificity}\tabularnewline
		\hline
		\textbf{BET} & $87.99\pm0.105$ & $79.68\pm0.123$ & $1.16\pm0.55$ & $80.76\pm0.123$ & $99.53\pm0.004$\tabularnewline
		\textbf{BSE} & $85.23\pm0.166$ & $77.17\pm0.206$ & $0.85\pm0.41$ & $78.90\pm0.216$ & $99.36\pm0.008$\tabularnewline
		\textbf{3DSS} & $91.42\pm0.133$ & $85.99\pm0.149$ & $0.72\pm0.22$ & $88.21\pm0.156$ & $99.1\pm0.01$\tabularnewline
		\textbf{ROBEX} & $94.32\pm0.081$ & $90.02\pm0.104$ & $0.61\pm0.24$ & $\mathbf{95.22\pm0.094}$ & $98.38\pm0.016$\tabularnewline
		\textbf{BEAST} & $90.00\pm0.165$ & $84.46\pm0.18$ & $0.61\pm0.15$ & $84.63\pm0.18$ & $\mathbf{99.93\pm0.001}$\tabularnewline
		\textbf{ASMCNN} & \textbf{$\mathbf{94.89\pm0.063}$} & $\mathbf{90.82\pm0.092}$ & $\mathbf{0.57\pm0.15}$ & $93.24\pm0.085$ & $99.36\pm0.007$\tabularnewline
		\hline
		\end{tabular}}
	}
\caption{2D Results of different methods in IBSR dataset.}
\label{tab:2DIBSR}
\end{table}

\begin{table}[!hbtp]
	\centering
	\centerline{
		\scalebox{0.9}{
		\begin{tabular}{llllll}
			\hline 
			\textbf{Method} & \textbf{Dice} & \textbf{Jaccard } & \textbf{Average Hausdorff} & \textbf{Sensitivity} & \textbf{Specificity}\tabularnewline
			\hline 
			\textbf{BET} & $93.25\pm0.084$ & $88.17\pm0.108$ & $0.72\pm0.28$ & $89.69\pm0.107$ & $99.34\pm0.006$\tabularnewline
			\textbf{BSE} & $94.37\pm0.103$ & $90.63\pm0.135$ & $0.57\pm0.32$ & $92.41\pm0.136$ & $99.33\pm0.007$\tabularnewline
			\textbf{3DSS} & $92.47\pm0.139$ & $87.93\pm0.153$ & $0.75\pm0.21	$ & $91.46\pm0.161$ & $98.22\pm0.016$\tabularnewline
			\textbf{ROBEX} & $93.85\pm0.127$ & $90.12\pm0.145$ & $0.62\pm0.17$ & $92.53\pm0.148$ & $99.08\pm0.006$\tabularnewline
			\textbf{BEAST} & $93.00\pm0.135$ & $88.83\pm0.154$ & $0.62\pm0.16$ & $89.65\pm0.154$ & $\mathbf{99.67\pm0.006}$\tabularnewline
			\textbf{ASMCNN} & \textbf{$\mathbf{95.39\pm0.07}$} & $\mathbf{91.87\pm0.103}$ & $\mathbf{0.57\pm0.19}$ & $\mathbf{94.72\pm0.07}$ & $99.33\pm0.01$\tabularnewline
			\hline 
		\end{tabular}
	}}
	\caption{2D Results of different methods in LPBA dataset.}
	\label{tab:2DLPBA}
\end{table}

\begin{table}[!hbtp]
	\centering
	\centerline{
		\scalebox{0.9}{
		\begin{tabular}{llllll}
			\hline 
			\textbf{Method} & \textbf{Dice} & \textbf{Jaccard } & \textbf{Average Hausdorff} & \textbf{Sensitivity} & \textbf{Specificity}\tabularnewline
			\hline 
			\textbf{BET} & $85.34\pm0.132$ & $76.17\pm0.154$ & $1.91\pm0.77$ & $78.17\pm0.158$ & $98.9\pm0.011$\tabularnewline
			\textbf{BSE} & $85.53\pm0.117$ & $76.18\pm0.145$ & $1.94\pm0.7$ & $78.45\pm0.15$ & $98.80\pm0.013$\tabularnewline 
			\textbf{3DSS} & $87.30\pm0.145$ & $79.64\pm0.173$ & $1.81\pm0.67$ & $82.91\pm0.178$ & $98.48\pm0.014$\tabularnewline
			\textbf{ROBEX} & $93.44\pm0.088$ & $88.59\pm0.112$ & $1.27\pm0.27$ & $92.1\pm0.104$ & $98.44\pm0.105$\tabularnewline
			\textbf{BEAST} & $88.56\pm0.142$ & $81.49\pm0.163$ & $1.59\pm0.37$ & $82.71\pm0.164$ & $\mathbf{99.26\pm0.001}$\tabularnewline
			\textbf{ASMCNN} & \textbf{$\mathbf{95.07\pm0.054}$} & $\mathbf{91.04\pm0.085}$ & $\mathbf{1.13\pm0.42}$ & $\mathbf{94.36\pm0.071}$ & $98.53\pm0.015$\tabularnewline
			\hline 
		\end{tabular}
	}}
	\caption{2D Results of different methods in OASIS dataset.}
	\label{tab:2DOASIS}
\end{table}

\begin{table}[!hbtp]
	\centering
	\centerline{
	\scalebox{0.9}{
		\begin{tabular}{llllll}
		\hline 
		\textbf{Method} & \textbf{Dice} & \textbf{Jaccard } & \textbf{Average Hausdorff} & \textbf{Sensitivity} & \textbf{Specificity}\tabularnewline
		\hline 
		\textbf{BET} & $89.63\pm0.035$ & $81.39\pm0.059$ & $34.99\pm6.57$ & $82.53\pm0.057$ & $99.57\pm0.002$\tabularnewline
		\textbf{BSE} & $88.48\pm0.09$ & $80.50\pm0.144$ & $32.06\pm11.76$ & $82.09\pm0.157$ & $99.42\pm0.005$\tabularnewline
		\textbf{3DSS} & $94.56\pm0.019$ & $89.75\pm0.035$ & $24.05\pm2.01$ & $92.02\pm0.048$ & $99.18\pm0.006$\tabularnewline
		\textbf{ROBEX} & $96.24\pm0.012$ & $92.78\pm0.021$ & $22.72\pm3.24$ & $\mathbf{97.14\pm0.015}$ & $98.47\pm0.012$\tabularnewline
		\textbf{BEAST} & $94.46\pm0.021$ & $89.57\pm0.037$ & $22.27\pm2.81$ & $89.76\pm0.038$ & $\mathbf{99.93\pm0.0004}$\tabularnewline
		\textbf{ASMCNN} & \textbf{$\mathbf{96.51\pm0.009}$} & $\mathbf{93.27\pm0.018}$ & $\mathbf{20.63\pm1.05}$ & $95.04\pm0.027$ & $99.39\pm0.004$\tabularnewline
		\hline 
		\end{tabular}
	}}

\caption{3D Results of different methods in IBSR dataset.}
\label{tab:3DIBSR}
\end{table}
\begin{table}[!hbtp]
	\centering
	\centerline{
	\scalebox{0.9}{
		\begin{tabular}{llllll}
		\hline 
		\textbf{Method} & \textbf{Dice} & \textbf{Jaccard } & \textbf{Average Hausdorff} & \textbf{Sensitivity} & \textbf{Specificity}\tabularnewline
		\hline 
		\textbf{BET} & $95.19\pm0.017$ & $90.87\pm0.031$ & $26.52\pm4.88$ & $92.33\pm0.032$ & $99.40\pm0.002$\tabularnewline
		\textbf{BSE} & $96.19\pm0.053$ & $93.09\pm0.086$ & $21.17\pm8.38$ & $94.66\pm0.091$ & $99.39\pm0.003$\tabularnewline
		\textbf{3DSS} & $95.85\pm0.005$ & $92.04\pm0.009$ & $24.52\pm2.34$ & $95.90\pm0.007$ & $98.41\pm0.006$\tabularnewline
		\textbf{ROBEX} & $96.91\pm0.002$ & $94.00\pm0.003$ & $21.88\pm1.02$ & $\mathbf{96.18\pm0.008}$ & $99.13\pm0.003$\tabularnewline
		\textbf{BEAST} & $96.47\pm0.006$ & $93.20\pm0.011$ & $21.11\pm1.99$ & $93.93\pm0.01$ & $\mathbf{99.72\pm0.004}$\tabularnewline
		\textbf{ASMCNN} & \textbf{$\mathbf{97.14\pm0.007}$} & $\mathbf{94.46\pm0.013}$ & $\mathbf{20.15\pm2.10}$ & $96.14\pm0.016$ & $99.33\pm0.003$\tabularnewline
		\hline 
		\end{tabular}
	}}
\caption{3D Results of different methods in LPBA dataset.}
\label{tab:3DLPBA}
\end{table}
\begin{table}[!hbtp]
	\centering
	\centerline{\scalebox{0.9}{
		\begin{tabular}{llllll}
		\hline 
		\textbf{Method} & \textbf{Dice} & \textbf{Jaccard } & \textbf{Average Hausdorff} & \textbf{Sensitivity} & \textbf{Specificity}\tabularnewline
		\hline 
		\textbf{BET} & $87.99\pm0.064$ & $79.09\pm0.095$ & $46.42\pm9.81$ & $80.73\pm0.103$ & $98.98\pm0.008$\tabularnewline
		\textbf{BSE} & $88.27\pm0.048$ & $79.32\pm0.075$ & $46.16\pm7.44$ & $81.15\pm0.087$ & $98.87\pm0.011$\tabularnewline
		\textbf{3DSS} & $90.26\pm0.07$ & $82.93\pm0.107$ & $40.54\pm10.32$ & $85.39\pm0.121$ & $98.57\pm0.01$\tabularnewline
		\textbf{ROBEX} & $95.64\pm0.008$ & $91.66\pm0.015$ & $30.31\pm1.93$ & $94.36\pm0.025$ & $98.49\pm0.008$\tabularnewline
		\textbf{BEAST} & $92.72\pm0.014$ & $86.45\pm0.025$ & $35.65\pm2.65$ & $87.59\pm0.03$ & $\mathbf{99.33\pm0.008}$\tabularnewline
		\textbf{ASMCNN} & \textbf{$\mathbf{96.50\pm0.009}$} & $\mathbf{93.25\pm0.016}$ & $\mathbf{26.17\pm2.23}$ & $\mathbf{95.82\pm0.019}$ & $98.58\pm0.006$\tabularnewline
		\hline 
		\end{tabular}
	}}
\caption{3D Results of different methods in OASIS dataset.}
\label{tab:3DOASIS}
\end{table}

As the proposed approach works with 2D scans, we formulate the segmented results on 3D structures by other algorithms as sequences of 2D images in the sagittal plane for an appropriate evaluation. 
BEAST has the highest mean of Specificity with a small standard deviation for three datasets, although there is no significant difference between six methods at this evaluation metric. From the result tables, we see that ASMCNN achieves the best performance with the highest average of Dice overlap (with 0.57 - 1.63$\%$ more upper than the second one), Jaccard index (0.8 - 2.45 $\%$ higher), and AHD on three datasets. The proposed approach gets the best Sensitivity on both LPBA and OASIS datasets, while ROBEX produces the highest on the IBSR dataset. Besides, BSE also achieves the same AHD score on the LPBA dataset similar to ASMCNN, but its standard deviation is higher than us (0.32 in comparison with 0.19). It turns out that ASMCNN has an outstanding performance in all evaluation metrics and gets the smallest standard deviation in the Dice coefficient, Jaccard index, and Sensitivity. Consequently, the experimental results illustrate the consistency of our method.  

Although processing with a 2D plane, we can construct 3D results by combining sequences of scans to make ASMCNN more comparable with other algorithms. There is an increase in the accuracy of five remaining methods in this type of evaluation with significant improvement at standard deviation. Among other algorithms, BEAST has major changes in its accuracy for several evaluated metrics on three datasets. For instance, it approximately gains more 3.47 - 5.1$\%$ on Dice coefficients, Jaccard indexes, and Sensitivity values. Furthermore, BEAST still provides the highest Specificity, which is almost 100$\%$ for all datasets.
Meanwhile, ROBEX continually obtains the best scores in Sensitivity on two datasets IBSE and LPBA. It is worth emphasizing that the 3D results produced by ASMCNN are remarkable and competitive with others. It preserves an impressive performance that gets the best scores on three datasets at the Dice coefficient, Jaccard index, AHD, and OASIS at Sensitivity.

The box-plots in Figure \ref{fig:IBSR_2d_Boxplot} and \ref{fig:IBSR_3d_Boxplot} visualize that BSE has a big interquartile range on IBSR which means the results given from it are more disperse and have a larger variance than other techniques. However, this method has the best performance on LPBA (Figure \ref{fig:LPBA_2d_Boxplot}, Figure \ref{fig:LPBA_3d_Boxplot}) which gives extremely accurate results and better than others although it has a terrible outlier (about 72$\%$ at Dice overlap for 2D evaluation). Indeed, this issue is mentioned above that BSE can provide very accurate segmentations and may give atrocious results based on tuning employed parameters. By contrast, ASMCNN shows it can be durable and unaffected by datasets that outperform than others with high accuracy and small interquartile range as well as the standard deviation on three datasets, especially on OASIS (Figure \ref{fig:OASIS_2d_Boxplot}, Figure \ref{fig:OASIS_3d_Boxplot}). Unfortunately, its Specificity is worse than several methods. However, it can be improved by employing a suitable post-processing method to mitigate several false positives in the segmentations. 

\begin{table}[!hbtp]
	\centering		
	\centerline{\scalebox{0.88}{
		\begin{tabular}{llll}
			\hline 
			\textbf{Method} & \textbf{Dice} & \textbf{Sensitivity} & \textbf{Specificity}\tabularnewline
			\hline 
			\textbf{2D-Unet (Dong et al. \cite{dong2017automatic})} & $94.36\pm0.03$ & $95.45\pm0.01$ & $96.87\pm0.01$
			\tabularnewline
			\textbf{3D-CNN (Kleesiek et al. (\cite{kleesiek2016deep})} & $95.19\pm0.01$ & $\mathbf{96.25\pm0.02}$ & $99.24\pm0.003$
			\tabularnewline
			\textbf{ASMCNN} & \textbf{$\mathbf{97.03\pm0.01}$} & $95.80\pm0.02$ & $\mathbf{99.38\pm0.01}$ 
			\tabularnewline
			\hline 
	\end{tabular}}}
\caption{3D combined results of different methods for IBSR, LPBA, and OASIS dataset.}
	\label{tab:3DCOMBINED}
\end{table}

\subsection*{Comparing with other deep learning methods}
As  results 3D-CNN in \cite{kleesiek2016deep} only have Dice, Sensitivity, and Specificity scores for IBSR, LPBA, and OASIS datasets,  we measured results of the 2D-Unet \cite{dong2017automatic} method on these three metrics. The results are shown in Table \ref{tab:3DCOMBINED}.

The experimental results indicate that our method achieved the highest scores on both Dice and Specificity with $97.03\%$ and $99.38\%$, respectively, while 3D-CNN gained the best value in Sensitivity at $96.25\%$.

\subsection{Ablation Study}
In this section, we analyze the contribution of the following factors: (i) the effectiveness of the combination of ASM with CNN to achieve prediction results, (ii) the advantages of the grouping mechanism to operate on three different groups rather than a single one and finally (iii) the contribution of CRF post-processing step. For (i), we experimented with two distinct cases: (i.a) training a single CNN-based model (Figure \ref{fig:struct_deep}) to predict for each pixel in all three groups and following by described post-processing steps; (i.b) utilized only ASM results without further refining by CNN to produce Group II and Group III predictions while remaining other components. For (ii), due to the large deformation of the Group I compared to other ones (Figures \ref{fig:group_all} and \ref{fig:three_axis}), the ASM could not learn a single principal shape simultaneously for all regions. We thus experimented by learning an ASM model for only Group II and Group III. All other steps, such as correcting with CNN or post-processing steps, were retained to highlight the grouping's role. Lastly, factor (iii) was checked by eliminating the CRF-step for both three regions to obtain final binary masks. 

We reported the average 3D results across all datasets for those settings in Table \ref{tab:3DCOMBINED_ABLATION}. One could observe that without grouping images, our performance declined dramatically from $97.03\%$ and $95.80\%$ down to $92.93\%$ and $88.12\%$ for Dice and Sensitivity, respectively. A similar trend also occurred without applying CNN or ASM, although it was lighter. Precisely, the Dice and Sensitivity metric decreased by $1.60\%/2.0\%$ and $3.07\%./ 2.04\%$ without CNN/ASM respectively. On the Specificity metric, our full settings also hold the best values than other options with the deviation from small ($0.25\%$, w/out CNN) to a large margin ( $3.49\%$, w/out ASM). This phenomenon showed that the CNN-based approach solely tends to recognize the location of skull tissues as brain tissues, resulting from the lack of global shape features from the ASM model. Lastly, the CRF component contributed significantly to the final performance, thereby improving Dice and Sensitivity from $95.69\%/94.63\%$ to $97.03\%/95.80\%$ correspondingly. Though there was a small reduction in the Specificity metric; however, the deviation was not significant.

\begin{table}[!hbtp]
	\centering		
	\centerline{\scalebox{0.88}{
		\begin{tabular}{llll}
			\hline 
			\textbf{Method} & \textbf{Dice} & \textbf{Sensitivity} & \textbf{Specificity}\tabularnewline
			\hline 
			\textbf{ASMCNN (Without ASM)} & $94.13\pm0.01$ & $93.76\pm0.01$ & $95.89\pm0.02$
						\tabularnewline
			\textbf{ASMCNN (Without CNN)} & $95.43\pm0.01$ & $92.73\pm0.02$ & $99.13\pm0.01$
			\tabularnewline
						\textbf{ASMCNN (Without Grouping)} & $92.93\pm0.03$ & $88.12\pm0.06$ & $98.26\pm0.01$
							\tabularnewline
			\textbf{ASMCNN (Without CRF)} & $95.69\pm0.01$ & $94.63\pm0.02$ & $\mathbf{99.41\pm0.01}$
			\tabularnewline
			\textbf{ASMCNN (Full Settings)} & \textbf{$\mathbf{97.03\pm0.01}$} & $\mathbf{95.80\pm0.02}$ & $99.38\pm0.01$ 
			\tabularnewline
			\hline 
	\end{tabular}}}
\caption{3D combined results of the ablation study for IBSR, LPBA, and OASIS dataset.}
	\label{tab:3DCOMBINED_ABLATION}
\end{table}

In short, all empirical pieces of evidence can be argued by the properties of employed algorithms. Firstly, without grouping shapes, our curve evolution based on ASM easily was stuck in a local optimum, leading to a massive discrepancy between the prediction and ground-truth targets, which is too difficult to improve by CNN. This phenomenon was also similar to the results of the BET algorithm depicted in Figures \ref{fig:IBSR_Compare} and \ref{fig:OASIS_Compare} (the top left pictures), in which the curve was stopped at the inner boundary inside the brain because of a confusion between this temporary state with a target shape learned from the training set. One of the causes for this confusion is that we trained a statistical shape model using patterns that differ significantly in shape and appearance, resulting in learned models that could not accurately capture the variation between regions. Secondly, utilizing ASM solely on separated regions may create over-smooth results; thereby, fine-grant details along the image boundary usually were skipped. In such situations, refining steps are necessary to compensate for this shortfall (Figure \ref{fig:group0203final}). In the opposite direction, the CNN-based strategy for each pixel using only local features could not consider important factors such as the consistency of brain shape and global appearances, implying incorrect predictions of skull-tissues with brain-tissues. Finally, the CRF post-processing steps allow further refining segmented results in terms of global consistency in appearance aspects, complementing for potential errors of CNN's local-based approaches.

\section{Discussion}
\paragraph{Our Strengths and Limitations}
We applied the divide-and-conquer strategy, which was inspired by exploiting brain regions' symmetric regions in the sagittal plane (Figure \ref{fig:three_axis}). By this, images belong to a particular group share similar attributes such as shapes and deformation along its boundaries, which was then leveraged to compose proper solutions to optimize final results.  It involves (i) capturing global shapes (by ASM) and refining details with local features (by CNN) for Groups II, III and (ii) learning a curve (by Gaussian Process) to adjust brain center points through connecting prior medical knowledge and current masks obtained from (i). In short, these mechanisms permit us a flexible way to control segmentation quality effectively in each group by taking into account desired characteristics during training and prediction procedure. 
Such properties are essential as dealing with small brain regions whose shapes are complex as Group I.  However, the proposed method also has some significant limitations. Firstly, ASMCNN requires more annotation efforts to train the whole system. In specific, besides labeled masks for each image (or labeled 3D volume) as default in segmentation task, we need further two kinds of data, including four image indexes (Algorithm \ref{alg:Training}) denoting for borderlines among groups and four ratio values for extracting bounding boxes which comprise specific features to highlight variations between disparate regions (Figure \ref{fig:group_hog}).  Secondly, our method also demands user to input some hyper-parameters, for instance in  \textit{CheckCenter} function (Algorithm \ref{alg:Check_Center}), \textit{CheckDistance} function (Algorithm \ref{alg:Check_Distance}) and in conditional random field post-processing. However, this requirement can also be considered an ASMCNN's feature since users can adjust the parameters to obtain the best results under different image acquisition conditions. Finally, for regions II and III, the CNN steps have to wait until the ASM procedure ending, or in Group I, we can only produce final predictions given obtained masks from Groups II and III. This sequential fashion results in our method being slower than the end-to-end deep network-based approach.

\paragraph{\textit{Comparing with Other Baselines}}
In the non-deep network category, ROBEX is the second-best method, following our ASMCNN in most metrics. This method runs directly on the 3D volume by evolving a point distribution model-based curve and could capture brain regions' s 3D global structure stably. However, in practice, the target brain shape is not perfectly represented; ROBEX thus usually ignores or over-segments small and complicated regions (Figure \ref{fig:OASIS_Compare} and \ref{fig:LPBA_Compare}, bottom right images). Simultaneously, with ASMCNN, we can overcome such challenges by adapting specific rules for Group I. To fix these harmful errors, ROBEX needs to improve further output contours driven by deep features as our approach with CNN. Unfortunately, training such networks is challenging due to the insufficient data if we work directly in 3D space instead of 2D slices. For instance, in our experiments, we have $103$ 3D volume samples for all datasets, compared to approximately $17000$ 2D images extracting from 3D volumes. Such a massive difference in the amount of data explains why a 2D based approach like ASMCNN could offer a greater benefit to training a dedicated deep network for correcting object boundaries. In the deep network category, 2D-Unet and 3D-CNN provide a convenient and efficient way to generate segmentation results by learning only labeled ground-truths. However, both methods predict binary masks by considering only local features; hence it is quite sensitive to noise regions whose gray levels closely resemble brain regions due to the image acquisition's errors. In contrast, our ASMCNN considers both global structure information by ASM and local features by utilizing the power of deep features in CNN, which implies a more robust and reliable pipeline.
  
\paragraph{\textit{Future Work}}
In our flowchart, ASM plays a role in preserving the principal shapes. However, ASM could only work effectively when the objects' shapes are similar to the trained active appearance model. Therefore, the algorithm may likely produce poorer results when processing unusual shapes. Unfortunately, in medical imaging, the data are usually obscured by noise due to limitations of image acquisition systems, or the shapes and the distribution of images are not included in training data. In such cases, ASM may produce severely imprecise boundaries; even CNN cannot verify or refine. We intend to investigate techniques based on deep learning to construct better models for shape and appearance in future work. In another direction, working on the whole 3D brain volume is also a promising approach. To this end, we can extend 3D-ASM \cite{heimann20053d} based methods so that prior medical knowledge, for example, symmetry property across two hemispheres, is considered during the curve evolution. Besides, learning optimal features along the object's profiles could be enhanced by leveraging deep features. Finally, modifying CNN-based refining strategies to make it more robust in 3D volume could be achieved by generalizing our architecture in Figure \ref{fig:struct_deep} from pixels to voxel input and combing transfer learning techniques.
\section{Conclusion}
This article has proposed a novel approach for brain extraction in magnetic resonance images, namely ASMCNN, by associating two main components: Active Shape Model and Convolutional Neural Network. Unlike existing methods, ASMCNN processes 2D sequences of images instead of using 3D brain structures. For each 2D scan, a rough estimation of the brain region is estimated by ASM. Its boundary is then refined by a CNN, which is constructed and trained with special rules. Finally, the brain region is post-processed by CRFs, Gaussian processes, and special rules. Our proposed approach has shown its consistency in performance; mainly, it can produce highly accurate segmentations in all cases, even when brain regions are small and scattered. In the experiments, our method has achieved remarkable for the whole three datasets (IBSR, LPBA, and OASIS) in 2D scans and 3D structures and surpassed the performance of seven other state-of-the-art methods.

\bibliography{mybibfile}
\clearpage
\section*{Appendices}
\label{sec:Appendices}
\begin{center}
	\begin{table}[htbp]  
		\begin{tabular}{|c|c|} 
			\hline
			\bf{Notation} & \bf{Remark} \\
			\hline
			$P=\{p_{1},p_{2},...,p_{m}\}$ & \makecell{The set of people in each dataset \\ where $p_{i}$ $(1\leq i\leq m)$ is the $i^{th}$ person.}\\
			\hline
			$N=\{n_{1},n_{2},...,n_{m}\}$ &\makecell{ $n_{j}$ is the number of 2D images \\for each person $p_j$  $(1\leq j\leq m)$.}\\ 
			\hline
			$I_{p_{j}}=\{I_{p_{j}1},I_{p_{j}2},...,I_{p_{j}n_{j}}\}$ &\makecell{ $I_{p_{j}k}$ $(1\leq k\leq n_{j})$ is the \\ $k^{th}$ image from person $p_{j}$.} \\
			\hline
			$G=\{G_{I},G_{II},G_{III}\}$ & \makecell{The list of groups of 2D images\\ determined by our algorithm.}\\
			\hline 
			\makecell{$G_{I-p_{j}}, G_{II-p_j}, G_{III-p_{j}}$} & \makecell{$G_{k-p_{j}}$ is the list of images in group \\$k$ ($k=I,II,III$)  for a person $p_j$ generated by our algorithm.}\\
			\hline
			$R_i,\ (i \in {1, 2, 3, 4})$ & \makecell{The vectors indicate for the starting and ending slices \\ for each group in training data.}\\
			\hline
			$F_i,\ (i \in \{I, II, III\})$ & \makecell{The feature vectors extracted by HOG algorithm \\ for images in group $G_{i}$}\\
			
 			\hline 
			$M_{12},\, M_{23}$ & \makecell{SVM models for image classification \\ between Group I vs. II and Group II vs. III.}\\
			\hline
			$r_{1},\, r_{2},\, r_{3},\, r_{4}$ & \makecell{All rates used for estimating the \\ starting and ending slices between different groups.}\\
			\hline     
		\end{tabular}
		\caption{List of notations used in Algorithms \ref{alg:Training} and \ref{alg:Testing}.}
		\label{table:t1}
	\end{table}
\end{center}

\begin{algorithm}
	\DontPrintSemicolon
	\SetAlgoLined
	\SetKwInOut{Input}{Input}\SetKwInOut{Output}{Output}
	\Input{$P$}
	\Output{$M_{12},\, M_{23},\, r_{1},\, r_{2},\, r_{3},\, _{4}$}
	\BlankLine
	\ForEach{person $p_{j}\in P$}{    
		select $k_{1},k_{2},k_{3},k_{4}$ such that:\\
		$A=\{I_{p_{j}1},I_{p_{j}2},...,I_{p_{j}k_{1}}\}\in G_{I-p_{j}}$
		
		$B=\{I_{p_{j}k_{1}+1},I_{p_{j}k_{1}+2},...,I_{p_{j}k_{2}}\}\in G_{II-p_{j}}$
		
		$C=\{I_{p_{j}k_{2}+1},I_{p_{j}k_{2}+2},...,I_{p_{j}k_{3}}\}\in G_{III-p_{j}}$
		
		$D=\{I_{p_{j}k_{3}+1},I_{p_{j}k_{3}+2},...,I_{p_{j}k_{4}}\}\in G_{II-p_{j}}$
		
		$E=\{I_{p_{j}k_{4}+1},I_{p_{j}k_{4}+2},...,I_{p_{j}n_{j}}\}\in G_{I-p_{j}}$
		
		insert:\\
		$G_{x-p_{j}}$ to vector $G_{x}\,(x\in{I, II, III})$ \\
		$k_{i}/n_{j}$ to vector $R_{i}\,(i\in{1,2,3,4})$ \, 
		\;
		
	}
	\BlankLine
	\ForEach{group  $G_{i}\in G$}{   
		select constant $r$\;
		\ForEach{image I $\in G_{i}$}{
			$Rec$  $\leftarrow$ Extract rectangle containing the skull of $I$ \\ using mask data\;
			$Sub Rec$ $\leftarrow$ Extract sub rectangle from $Rec$ with $r$\;
			$feature\,\leftarrow$\,Extract features for $I$ in $SubRec$ with \\ $HOG$ method \cite{dalal2005histograms}\;
			insert $feature$ to vector $F_{i}$
		}
	}
	
	$M_{12} \leftarrow $ Training a classifier model with $F_{1}$, and $F_{2}$ by soft-margin SVM  \;
	$M_{23} \leftarrow $ Training a classifier model with $F_{2}$, and $F_{3}$ by  soft-margin SVM   \;
	$r_{i} \leftarrow mean\,(R_{i});\  i\in{1,2,3,4})$\;
	\Return $M_{12},\, M_{23},\, r_{1},\, r_{2},\, r_{3},\, r_{4}$
	
	\caption{Training classifiers for clustering images into different groups}
	\label{alg:Training}
\end{algorithm}

\begin{algorithm}
	\DontPrintSemicolon
	\SetAlgoLined
	\SetKwInOut{Input}{Input}\SetKwInOut{Output}{Output}
	\Input{$M_{12},\, M_{23},\, r_{1},\, r_{2},\, r_{3},\, r_{4},\, I_{p_{j}}$}
	\Output{$G_{I-p_{j}},\, G_{II-p_{j}},\, G_{III-p_{j}}$}
	\BlankLine
	$r_{i\, new}\leftarrow r_{i}.n_{j}\ (i\in\{1,2,3,4\})$, \,$n_{j}$ is the number images of $p_{j}$
	\;
	\For{$id\leftarrow r_{1,\, new}-10$ to $r_{1,\, new}+10$}{    
		\If{$M_{12}(I_{p_{j}id})==0$}{
			insert $A=\{I_{p_{j}1},I_{p_{j}2},...,I_{p_{j}id-1}\}$ to vector $G_{I-p_{j}}$\;
			$r_{1,\, new}\leftarrow id-1$ \;
			break\;
		}
		
	}
	\BlankLine
	\For{$id\leftarrow r_{2,\, new}-10$ to $r_{2,\, new}+10$}{    
		\If{$M_{23}(I_{p_{j}id})==0$}{
			insert $B=\{I_{p_{j}r_{1,new}+1},I_{p_{j}r_{1,new}+2},...,I_{p_{j}id-1}\}$\,to vector $G_{II-p_{j}}$\;
			$r_{2,\, new}\leftarrow id-1$ \;
			break\;
		}
		
	}
	
	\BlankLine
	\For{$id\leftarrow r_{3,\, new}-10$ to $r_{3,\, new}+10$}{    
		\If{$M_{23}(I_{p_{j}id})==1$}{
			insert $C=\{I_{p_{j}r_{2,new}+1},I_{p_{j}r_{2,new}+2},...,I_{p_{j}id-1}\}$\,to vector $G_{III-p_{j}}$\;
			$r_{3,\, new}\leftarrow id-1$ \;
			break\;
		}
		
	}
	
	\BlankLine
	\For{$id\leftarrow r_{4,\, new}-10$ to $r_{4,\, new}+10$}{    
		\If{$M_{12}(I_{p_{j}id})==1$}{
			insert $D=\{I_{p_{j}r_{3,new}+1},I_{p_{j}r_{3,new}+2},...,I_{p_{j}id-1}\}$\,to vector $G_{II-p_{j}}$\;
			$r_{4,\, new}\leftarrow id-1$ \;
			insert
			$E=\{I_{p_{j}r_{4,new}+1},I_{p_{j}r_{4,new}+2},...,I_{p_{j}n_{j}}\}$\,to vector $G_{I-p_{j}}$\;
			break\;
		}
		
	}
	\Return $G_{I-p_{j}},\, G_{II-p_{j}},\, G_{III-p_{j}}$
	
	\caption{Classifying images into three groups $G_{I},\, G_{II},\,G_{III}$}
	\label{alg:Testing}
\end{algorithm}

\begin{algorithm}
	\DontPrintSemicolon
	\SetAlgoLined
	\SetKwInOut{Input}{Input}\SetKwInOut{Output}{Output}
	\Input{$A, \, B, \, C, \,D, \,x_{AB}$}
	\Output{$x_{CD}$}
	\BlankLine
	$x_{CD} = C + (x_{AB} - A)*(D - C)/(B - A)$
	\BlankLine
	\Return $x_{CD}$
	\caption{\textit{ConvertRange} --  Convert a value $x_{AB}$ in range $[A, B]$ to range $[C, D]$}
	\label{alg:Convert_Range}
\end{algorithm}

\begin{algorithm}
	\DontPrintSemicolon
	\SetAlgoLined
	\SetKwInOut{Input}{Input}\SetKwInOut{Output}{Output}
	\Input{$C_{p_j\,1} \rightarrow C_{p_jm}, C_{p_j \,k+1} \rightarrow C_{p_j\,n_j}$ : CNN resulting images of the $p_j$  person in Group 01 from the slice $1^{th}  \rightarrow m^{th}$ and $(k+1)^{th}  \rightarrow n_j^{th}$. \\ $F_{p_j\,m+1}, \,F_{p_j\,k}$ : two final segmentation results of the $p_j$ \;person in Group II at slice $(m+1)^{th}$ and $k^{th}$.\;$GPM$ : Gaussian Process Model is described in subsection\ref{sec:group-i}.}
	\Output{Final predictions for the $p_j$ person in Group I $F_{p_j\,1} \rightarrow F_{p_jm}, F_{p_j \,k+1} \rightarrow F_{p_j\,n_j}$}
	\BlankLine
	$Image_1 = F_{p_j\,m+1}$, $Image_2 = F_{p_j\,k}$\;
	$\alpha = 0.4$, $\beta = 1.75$\;
	\BlankLine
	\For{$i \leftarrow  m\  \text{down to}\ \  1$}
	{
		$C_{p_j\,i} \leftarrow$ Denoise in $C_{p_j,\,i}$\;
		$C_{p_j\,i} = CheckCenter(C_{p_j\,i},Image_1)$\;
		$C_{p_j\,i} = CheckArea({p_j\,i},\alpha)$\;
		$d_i \leftarrow GPM(i_{GPM})$\;
		$C_{p_j\,i} = CheckDistance({p_j\,i}, d_i, \beta)$\; 
		$F_{p_j\,i} \leftarrow CRF(C_{p_j\,i})$\;
		$Image_1 \leftarrow F_{p_j\,i}$\;
	}
	\BlankLine
	\For{$i \leftarrow k+1\ \text{to}\ \ n_j$}
	{
		$C_{p_j\, i} \leftarrow$ Denoise in $C_{p_j\, i}$\;
		$C_{p_j\, i} = CheckCenter(C_{p_j\, i} ,Image_2)$\;
		$C_{p_j\, i} = CheckArea(C_{p_j\, i},\alpha)$\;
		$d_i \leftarrow GPM(i_{GPM})$\;
		$C_{p_j\, i} = CheckDistance(C_{p_j\, i}, d_i, \beta)$\;
		$F_{p_j\,i} \leftarrow CRF(C_{p_j\, i})$\;
		$Image_2 = F_{p_j\,i}$\;
	}
	\BlankLine
	\Return $F_{p_j\,1} \rightarrow F_{p_jm}, F_{p_j \,k+1} \rightarrow F_{p_j\,n_j}$
	\caption{Process Images in Group I for the $p_j$ person }
	\label{alg:PIG1}
\end{algorithm}

\begin{algorithm}
  	\DontPrintSemicolon
  	\SetAlgoLined
  	\SetKwInOut{Input}{Input}\SetKwInOut{Output}{Output}
  	\Input{Two binary images $X,\,Y;\;$ Open distance $\alpha$
  	}
  	\Output{
  		Binary image $Z$ created by combining $X$ and $Y$
  	}
  	\BlankLine
  	$O \leftarrow OpenBoundary(Y,\alpha)$\;
  	$Core \leftarrow Y\backslash O$ and $Boundary \leftarrow X \cap O$\;
  	$Z \leftarrow Core \cup Boundary$\;
  	Apply morphology to fill small holes in $Z$ (if exists)
  	
  	\BlankLine
  	\Return $Z$
  	
  	\caption{$MergeSlice(X,\,Y,\,\alpha)$}
  	\label{alg:MergeSlice}
\end{algorithm}


\begin{algorithm}
	\DontPrintSemicolon
	\SetAlgoLined
	\SetKwInOut{Input}{Input}\SetKwInOut{Output}{Output}
	\Input{$X, \,Y$}
	\Output{$Z$}
	\BlankLine
	\BlankLine
	$Z = X$
	\BlankLine
	$bbox \leftarrow$ the smallest bounding box surrounding all components in Y
	\BlankLine
	\ForEach{component $c_i$ in $Z$}
	{
		$center_i \leftarrow$ Coordinates of the center of $c_i$\\
		\If{$center_i$ is not in $bbox$}
		{Remove $c_i$ out of $Z$}
	}
	\BlankLine
	\Return $Z$
	\caption{\textit{CheckCenter} -- 	Remove components in image $X$ whose center is not in image $Y$ to create image $Z$}
	\label{alg:Check_Center}
\end{algorithm}

\begin{algorithm}
	\DontPrintSemicolon
	\SetAlgoLined
	\SetKwInOut{Input}{Input}\SetKwInOut{Output}{Output}
	\Input{$X, \,\alpha$}
	\Output{$Y$}
	\BlankLine
	$Y = X$
	\BlankLine
	\ForEach{component $c_i$ in $Y$}
	{
		$a_i \leftarrow$ Area of $c_i$
	}
	\BlankLine
	$a_{max} \leftarrow$ \textit{max} $\{a_i\}$\\
	$a_{threshold} = \alpha*a_{max}$
	\BlankLine
	\ForEach{component $c_i$ in $Y$}
	{
		\If{$a_i < a_{threshold}$}{Remove $c_i$ out of $Y$} 
	}
	\BlankLine
	\Return $Y$
	\caption{\textit{CheckArea} -- Remove small components in image $X$ with the coefficient $\alpha$ to create image $Y$}
	\label{alg:Check_Area}
\end{algorithm}

\begin{algorithm}
	\DontPrintSemicolon
	\SetAlgoLined
	\SetKwInOut{Input}{Input}\SetKwInOut{Output}{Output}
	\Input{$X, \,d, \, \beta$}
	\Output{$Y$}
	\BlankLine
	$Y = X$
	\BlankLine
	\ForEach{component $c_i$ in $Y$}
	{
		$center_i \leftarrow$ Coordinates of the center of $c_i$\\
		$dif_i = |d - \|center_i\|_2|$
	}
	\BlankLine
	$dif_{min} \leftarrow$ \textit{min} $\{dif_i\}$\\
	$dif_{threshold} = \beta*dif_{min}$
	\BlankLine
	\ForEach{component $c_i$ in $Y$}
	{
		\If{$dif_i > dif_{threshold}$}{Remove $c_i$ out of $Y$} 
	}
	\BlankLine
	\Return $Y$
	\caption{\textit{CheckDistance} -- Remove components having small distance in image $X$ with the coefficient $\beta$ to create image $Y$}
	\label{alg:Check_Distance}
\end{algorithm}

\clearpage

\end{document}